%% file: main.tex
\documentclass[11pt,table]{article}
\pdfoutput=1

\usepackage[utf8]{inputenc} 
\usepackage[T1]{fontenc}    
\usepackage[in]{fullpage}
\usepackage[dvipsnames]{xcolor}         
\usepackage{booktabs}       
\usepackage{amsfonts}       
\usepackage{nicefrac}       
\usepackage{microtype}      
\usepackage{amsmath,amssymb,amsthm}
\usepackage{mathtools}
\usepackage{mathrsfs}
\usepackage{thmtools}
\usepackage{thm-restate}
\usepackage{comment}
\usepackage{parskip}
\usepackage{siunitx}
\usepackage{etoolbox}
\usepackage{enumitem}
\usepackage{graphicx}
\usepackage{tabularx}
\usepackage[numbers,sort]{natbib}
\usepackage[hidelinks]{hyperref}
\usepackage{longtable}
\usepackage{afterpage}
\usepackage{etoc}
\usepackage{xurl}

\usepackage{url}            
\usepackage{microtype}      
\usepackage{enumitem}       
\usepackage{subfig}         
\usepackage{graphicx}       
\usepackage{multirow}       
\usepackage{floatrow}       

\definecolor{light-light-gray}{gray}{0.90} 

\extrafloats{100}

\newtoggle{showtodos}
\togglefalse{showtodos}

\newtoggle{showappendix}
\toggletrue{showappendix}

\newtoggle{isicml}
\togglefalse{isicml}

\input{latexdefs}

\newfloatcommand{capbtabbox}{table}[][\FBwidth]

\def\@fnsymbol#1{\ensuremath{\ifcase#1\or *\or \dagger\or \ddagger\or
   \mathsection\or \mathparagraph\or \|\or **\or \dagger\dagger
   \or \ddagger\ddagger \else\@ctrerr\fi}}
\newcommand{\ssymbol}[1]{^{\@fnsymbol{#1}}}

\title{LAION-5B: An open large-scale dataset for training next generation image-text models}

\author{
\textbf{Christoph Schuhmann}$^1$ §§°° \quad \textbf{Romain Beaumont$^1$} §§°° \quad \textbf{Richard Vencu}$^{1,3,8}$ §§°° \\ \textbf{Cade Gordon}$^{2}$ §§°°  
\quad \textbf{Ross Wightman}$^1$§§ \quad \textbf{Mehdi Cherti} $^{1,10}$§§ \\ 
\textbf{Theo Coombes$^1$} \quad \textbf{Aarush Katta$^1$} \quad \textbf{Clayton Mullis$^1$} \quad \textbf{Mitchell Wortsman$^{6}$}  \\
\textbf{Patrick Schramowski}$^{1,4,5}$ \quad \textbf{Srivatsa Kundurthy}$^{1}$  \quad \textbf{Katherine Crowson$^{1,8,9}$}  \\ 
 \quad  \textbf{Ludwig Schmidt}$^6$ °° \quad \textbf{Robert Kaczmarczyk}$^{1,7}$ °° \quad \textbf{Jenia Jitsev}$^{1,10}$ °° \\
LAION$^1$ \quad UC Berkeley$^2$ \quad Gentec Data$^3$ \quad TU Darmstadt$^4$ \quad Hessian.AI$^5$\\
University of Washington, Seattle$^6$ \quad
Technical University of Munich$^7$ \quad Stability AI$^8$ \\ 
EleutherAI$^9$ \quad Juelich Supercomputing Center (JSC), Research Center Juelich (FZJ)$^{10}$ \\
\texttt{contact@laion.ai} \\
\texttt{§§ Equal first contributions, °° Equal senior contributions}
}

\begin{document}

\date{}

\maketitle

\begin{abstract}
Groundbreaking language-vision architectures like CLIP and DALL-E proved the utility of training on large amounts of noisy image-text data, without relying on expensive accurate labels used in standard vision unimodal supervised learning. The resulting models showed capabilities of strong text-guided image generation and transfer to downstream tasks, while performing remarkably at zero-shot classification with noteworthy out-of-distribution robustness. Since then, large-scale language-vision models like ALIGN, BASIC, GLIDE, Flamingo and Imagen made further improvements. Studying the training and capabilities of such models requires datasets containing billions of image-text pairs. Until now, no datasets of this size have been made openly available for the broader research community. To address this problem and democratize research on large-scale multi-modal models, we present LAION-5B - a dataset consisting of 5.85 billion CLIP-filtered image-text pairs, of which 2.32B contain English language. We show successful replication and fine-tuning of foundational models like CLIP, GLIDE and Stable Diffusion using the dataset, and discuss further experiments enabled with an openly available dataset of this scale. Additionally we provide several nearest neighbor indices, an improved web-interface for dataset exploration and subset generation, and detection scores for watermark, NSFW, and toxic content detection. \footnote{Project page:  \href{https://laion.ai/laion-5b-a-new-era-of-open-large-scale-multi-modal-datasets/}{https://laion.ai/laion-5b-a-new-era-of-open-large-scale-multi-modal-datasets/}}
\end{abstract}

\etocdepthtag.toc{mtsection}



\section{Introduction}
\label{sec:intro}
\input{sections/introduction}

\section{Related Work}
\label{sec:related_works}

\input{sections/related_work}

\section{Collection Methodology}
\label{sec:collection_methodology}
\input{sections/collection_methods}

\section{Dataset Composition}
\label{sec:dataset_composition}
\input{sections/composition}

\section{Experiments Validating LAION-5B}
\label{sec:experiments}
\input{sections/experiments}

\section{Technical Limitations}
\label{sec:strengths_and_weaknesses}
\input{sections/strengths_and_weaknesses}

\section{Safety and Ethical Discussion}
\label{sec:ethical_discussion}
\input{sections/ethical_social}

\section{Conclusion}
\label{sec:conclusion}
\input{sections/conclusion}



\section*{Acknowledgments}
\input{sections/acknowledgments}

\input{sections/references}

\newpage


\appendix
\section*{Appendix (LAION-5B: An open large-scale dataset for training next generation image-text models)}

\section{Datasheet for LAION-5B dataset}
\label{sec:datasheet}
\input{sections/datasheet}

\section{Dataset Setup Procedure}
\input{sections/dataset_preparation}

\section{Dataset Preparation and Curation Details}
\label{sec:appendix_preparation_curation}
\input{sections/technical_curation}

\section{Dataset Samples and Statistics}
\input{sections/experiment_samples}


\section{Further Experimental Details and Results on CLIP reproduction}
\label{sec:appendix_experiments}
\input{sections/experiments_appendix}

\section{Overview of Experiments and Results on Generative Models}
\label{sec:appendix_experiments_generative}
\input{sections/experiments_generative_appendix}

\section{Further Discussion on Safety and Ethics}
\label{sec:appendix_safety}
\input{sections/safety_appendix}

\clearpage
\section*{Author contributions}
\input{sections/author_contributions}

\section*{Acknowledgments details}
\input{sections/acknowledgments_appendix}




%
%
%
%
%
%
%
%
%
%
%
%
%
%

\end{document}

%% file: latexdefs.tex
\newcommand{\Eop}{\ensuremath{\mathop{\mathbb{E}}}}
\newcommand{\E}{\Eop}
\newcommand{\R}{\ensuremath{\mathbb{R}}}
\newcommand{\N}{\ensuremath{\mathcal{N}}}

\newcommand{\D}[1]{\ensuremath{D_{\text{#1}}}}

\newcolumntype{L}[1]{>{\raggedright\arraybackslash}p{#1}}
\newcolumntype{C}[1]{>{\centering\arraybackslash}p{#1}}
\newcolumntype{R}[1]{>{\raggedleft\arraybackslash}p{#1}}

\iftoggle{showtodos}{
\newcommand{\becca}[1]{\todo[color=green!40]{Becca: #1}}
\newcommand{\ludwig}[1]{\todo[color=blue!40]{Ludwig: #1}}
}{
\newcommand{\becca}[1]{}
\newcommand{\ludwig}[1]{}
}

%% file: sections/introduction.tex


\label{intro}

Learning from multimodal data such as text, images, and audio is a longstanding research challenge in machine learning \citep{mori1999image, quattoni2007learning, weston2010large,karpathy2015deep, xu2015show}. 
Recently, contrastive loss functions combined with large neural networks have led to breakthroughs in the generalization capabilities of vision and language models \citep{radford2021learning, DALL-E, imagen}.
For instance, OpenAI's CLIP models \citep{radford2021learning} achieved large gains in zero-shot classification on ImageNet \citep{ILSVRC15}, improving from the prior top-1 accuracy of 11.5\%  \citep{li2017learning} to 76.2\%.
In addition, CLIP achieved unprecedented performance gains on multiple challenging distribution shifts~\citep{taori2020measuring,pmlr-v97-recht19a,imagenetr, imagenetsketch,objectnet,vidrobust}.
Inspired by CLIP’s performance, numerous groups have further improved image-text models by increasing the amount of computation and the training set size \citep{ALIGN, basic, zhai2021lit,yu2022coca}. 
Another recent success of multimodal learning is in image generation, where DALL-E \citep{DALL-E} and later models \citep{DALLE-2,imagen, GLIDE, parti, rombach2021highresolution} demonstrated the potential of text-guided image generation by producing high-quality images specific to the provided text. 

A critical ingredient in this new generation of image-text models is the pre-training dataset.
All of the aforementioned advances rely on large datasets containing hundreds of millions or even billions of image-text pairs, e.g., 400 million for CLIP \citep{radford2021learning} and 6.6 billion for BASIC \citep{basic}.
However, \emph{none of these datasets are publicly available}.
While OpenAI still released the CLIP models publicly \citep{radford2021learning}, later papers made neither the pre-training dataset nor the resulting models available to the wider research community \citep{ALIGN,basic,alayrac2022flamingo,parti,imagen,GLIDE,yu2022coca}.
As a result, research in this area has pooled into a small number of industrial research labs, limiting transparency and impeding research progress.
%

\begin{figure}[htb]
\begin{floatrow}
\ffigbox{%
  \centering
  \includegraphics[width=0.48\textwidth]{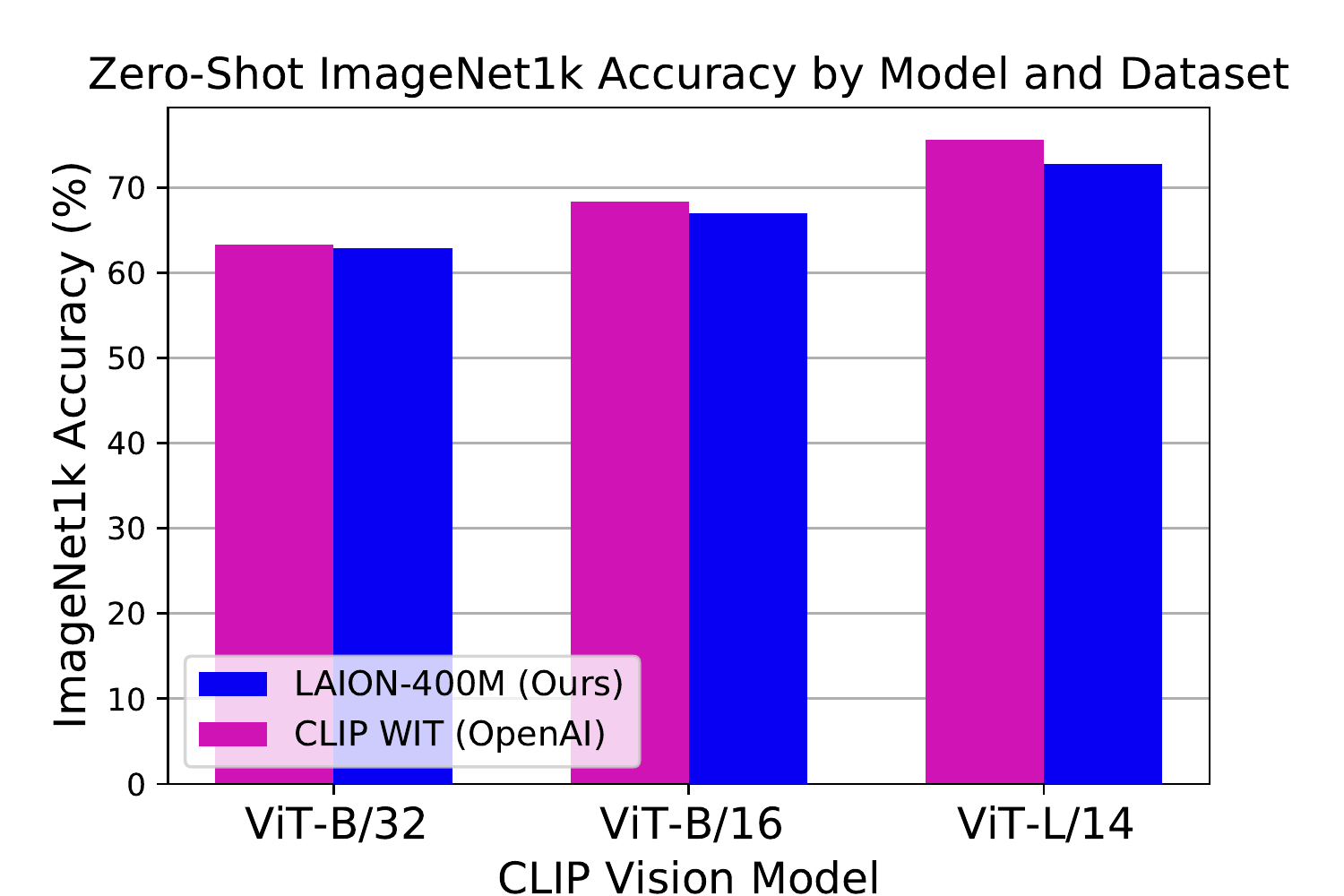}%
}{%
  \caption{\textbf{Zero-Shot Accuracy.} CLIP models trained on LAION-400M (ours) \cite{schuhmann2021laion}, a previously released subset of LAION-5B, show competitive zero-shot accuracy compared to CLIP models trained on OpenAI's original training set WIT when evaluated on ImageNet1k.}%
  \label{fig:sub1}
}
\capbtabbox{%
\centering
        \begin{footnotesize}
        \rowcolors{2}{light-light-gray}{white}
        \begin{tabular}{cc}\hline
        \textbf{Dataset}  & \textbf{\# English Img-Txt Pairs} \\
        \hline
        \multicolumn{2}{c}{\textbf{Public Datasets}}         \\ \hline
        MS-COCO  & 330K                    \\
        CC3M     & 3M                      \\
        Visual Genome & 5.4M               \\
        WIT      & 5.5M                    \\
        CC12M    & 12M                     \\
        RedCaps  & 12M                     \\
        YFCC100M & 100M\footnotemark \\
        \textbf{LAION-5B (Ours)} & \textbf{2.3B}                    \\ \hline
        \multicolumn{2}{c}{\textbf{Private Datasets}}         \\ \hline
        CLIP WIT (OpenAI) & 400M                    \\
        ALIGN    & 1.8B                    \\
        BASIC    & 6.6B                      \\ \hline
        \end{tabular}
        \end{footnotesize}
}{%
  \caption{\textbf{Dataset Size.} LAION-5B is more than 20 times larger than other public English image-text datasets.
  We extend the analysis from \citet{desai2021redcaps} and compare the sizes of public and private image-text datasets.}%
  \label{fig:sub2}
}
\end{floatrow}
\end{figure}

In this work, we address this challenge and make multimodal training more accessible by assembling a public dataset that is suitable for training large image-text models.
Specifically, we introduce LAION-5B, the largest public image-text dataset containing over 5.8 billion examples (see Table \ref{fig:sub2} for a comparison).
By starting from Common Crawl \citep{common_crawl} and filtering this data source with an existing CLIP model, we derive a dataset consisting of three parts: 2.32 billion English image-text examples, 2.26 billion multilingual examples, and 1.27 billion examples that are not specific to a particular language (e.g., places, products, etc.).
Beyond assembling the dataset, we also explore its ethical implications and flaws that emerge with large-scale data collection.
By releasing LAION-5B publicly, we offer the first opportunity for the community to audit and refine a dataset of this magnitude. 


\footnotetext{Although YFCC100M contains 100M image-text pairs, it is unclear how well the text matches the image for an average example from the dataset. \citet{CLIP}'s curation procedure reduced YFCC100M to 15M samples.}

To validate that LAION-5B is indeed suitable for training large image-text models, we conduct multiple experiments.
We focus on matching the performance of OpenAI's CLIP models because they are the largest publicly released image-text models.
OpenAI's CLIP models were trained on 400~million image-text pairs, and hence we also train CLIP models on a subset of LAION-5B containing the same number of examples (``LAION-400M'').
Across a diverse range of problem settings including ImageNet (zero-shot), distribution shifts, VTAB, retrieval, and fine-tuning, our models trained on LAION-400M match or come close to the performance of OpenAI's CLIP models.
Our ViT-L/14 models trained with OpenCLIP are the first open source reproductions of the largest CLIP models released by OpenAI.

Despite these validation results, LAION-5B is \emph{not} a finished data product.
Due to the immense size of current image-text pre-training datasets, curating LAION-5B for widespread use goes beyond the scope of a single research paper.
Hence we do not only release our dataset, but also our software stack we built for assembling LAION-5B.
We view our initial data release and this paper as a first step on the way towards a widely applicable pre-training dataset for multimodal models.
As a result, \textbf{we strongly recommend that LAION-5B should only be used for academic research purposes in its current form.
We advise against any applications in deployed systems without carefully investigating behavior and possible biases of models trained on LAION-5B.}

The remainder of the paper proceeds as follows.
After reviewing related work, we present our data collection process for LAION-5B in Section~\ref{sec:collection_methodology}.
Section~\ref{sec:dataset_composition} then describes LAION-5B's composition including its various subsets.
To validate LAION-5B, we reproduce and evaluate different image-text models in Section~\ref{sec:experiments}.
Before concluding, we discuss the technical limitations of LAION-5B in Section~\ref{sec:strengths_and_weaknesses} and safety and ethics concerns in Section~\ref{sec:ethical_discussion}.

%% file: sections/related_work.tex
\textbf{Vision-Language Models.} \citet{radford2021learning} made a large step forward in multimodal learning for image-text data with their CLIP (Contrastive Language–Image Pre-training) model.
The authors proposed a contrastive learning scheme to embed both images and text into a shared representation space, which enabled unparalleled performance in zero-shot image classification.
Moreover, CLIP made large progress on multiple challenging distribution shifts \citep{taori2020measuring, wortsman2021robust}.

After CLIP's initial success, ALIGN and BASIC improved contrastive multimodal learning by increasing the training set size and the batch size used for training \citep{ALIGN, basic}.
LiT also increased training scale and experimented with a combination of pre-trained image representations and contrastive fine-tuning to connect frozen image representations to text \citep{zhai2021lit}.
Flamingo introduced the first large vision-language model with in-context learning \citep{alayrac2022flamingo}.
Other papers have combined contrastive losses with image captioning  to further improve performance \citep{yu2022coca,li2022blip}.
Beyond image classification and retrieval, the community later adapted CLIP to further vision tasks such as object navigation and visual question answering \citep{mokady2021clipcap, khandelwal2021simple, shen2021much,cow}.

Another direction that has recently seen large progress in multimodal learning is text-guided image generation \citep{mansimov2015generating, reed2016generative, zhang2017stackgan}.
Specifically, DALL-E demonstrated diverse image generation capabilities for text prompts combining multiple concepts \citep{DALL-E}. 
GLIDE, DALL-E 2, Imagen, Parti, and Stable Diffusion then improved visual fidelity and text-prompt correspondence \citep{GLIDE, imagen, DALLE-2, parti, rombach2021highresolution}.

\textbf{Image-Text Datasets.} Earlier dataset creation efforts such as MS-COCO and Visual Genome curated image and region labels through human annotation \citep{lin2014microsoft, krishna2017visual}.
While this resulted in high-quality labels, it also limited the scale of the datasets to only 330K and 5M examples, respectively.
The web-harvested YFCC-100M dataset is substantially larger with about 99 million images and one million videos from Flickr, but only contains the user-generated metadata without additional annotations collected specifically for training computer vision models \citep{thomee2016yfcc100m}.
As a result, the text associated with an image sometimes has little to no correspondence with the actual image content.

To address this shortcoming of web-harvested image-text data, the Conceptual Captions dataset (CC3M) started with images and alt-text collected from the web, but then performed additional data cleaning procedures \citep{sharma-etal-2018-conceptual}.
To increase the size of the dataset, researchers later relaxed the filtering protocol to arrive at the subsequent CC12M dataset \citep{changpinyo2021conceptual}.
Building datasets from alt-text continued with ALT200M \citep{hu2021scaling} and ALIGN \citep{ALIGN}, which increased the dataset size up to 1.8 billion image-text pairs.
In contrast to relying on alt-text, RedCaps used the captions provided by Reddit users to collect higher quality captions \citep{desai2021redcaps}.

Datasets with non-English image-text pairs are less common.
As a result, researchers translated English captioning datasets to other languages such as Farsi, Korean, and Japanese \citep{ko-clip, japanese-clip, clip-fa}. 
To the best of our knowledge, the largest multilingual dataset before LAION-5B has around 36 million samples from Wikipedia Image Text \citep{wit}.
With the release of LAION-5B, researchers now have access to roughly two orders of magnitude more multilingual samples, which provides new opportunities for research on low-resource languages and multilingual models.

\textbf{Scaling Behavior.} Improving model performance by increasing data scale has been a theme in machine learning since at least the ImageNet dataset \citep{deng2009imagenet}.
In the following decade, computer vision benefited from growth in model, data, and compute scale, in addition to advances in both convolutional and transformer architectures \citep{kolesnikov2020big, dosovitskiy2020image, vaswani2017attention, xiaohua2021scaling}.
Industrial research labs assembled large internal datasets such as Instagram-1B, JFT300M, and JFT3B to support image pre-training \citep{mahajan2018exploring, sun2017revisiting, zhai2021scaling}.
Natural language processing (NLP) demonstrated the beneficial effect of model, data, and compute scale on generalization through large language models such as GPT-3 \citep{brown2020language} and associated experiments on scaling behavior \citep{kaplan2020scaling}.
Community efforts like the The Pile \citep{gao2020pile} and BigScience ROOTS \citep{bigscience_roots} made large text datasets more accessible.


%
%
%
%
%
%
%

%% file: sections/collection_methods.tex

We constructed LAION-5B starting from Common Crawl, a public web archive \citep{common_crawl}.
The Common Crawl organization crawls the web since 2008 and publishes the results in snapshots approximately every month.
Recent snapshots each contain about 300 TiB of data for around 3 billion web pages.
In the following, we introduce our pipeline to assemble and filter a vision-language dataset from images in Common Crawl and their associated HTML alt-text.

\subsection{Dataset Assembly Pipeline}
\begin{figure}[!t]
    \centering
    \includegraphics[width=0.95\textwidth]{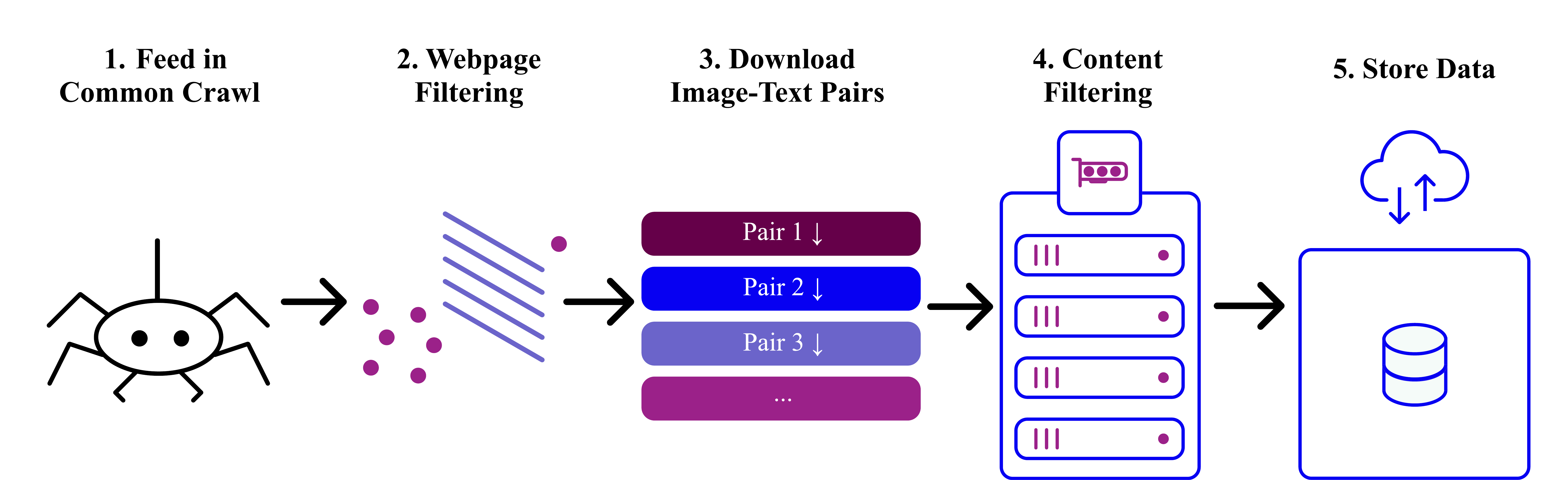}
    \caption{\textbf{Overview of the acquisition pipeline:} Files are downloaded, tracked, and undergo distributed inference to determine inclusion. Those above the specified CLIP threshold are saved.}
    \label{arch}
\end{figure}

Our dataset assembly pipeline follows the flowchart of Figure \ref{arch}.
At a high level, the pipeline consists of three main components: 
(i) distributed filtering of the Common Crawl web pages, 
(ii) distributed downloading of image-text pairs,
and (iii) content filtering.
The code used for the dataset pipeline may be found on GitHub\footnote{\url{https://github.com/rvencu/crawlingathome-gpu-hcloud}}.
We now describe each component in more detail. 

\textbf{Web page filtering.}
To extract image-text pairs from Common Crawl, we parse the HTML \texttt{IMG} (image) tags from Common Crawl's WAT metadata files.\footnote{See \url{https://commoncrawl.org/the-data/get-started/} for details of the metadata format.}
Specifically, we focus on images with an \emph{alt-text} so we can create image-text pairs.
The alt-text is an HTML attribute of \texttt{IMG} tags containing alternative text for situations where the corresponding image cannot be rendered.
For instance, screen reader software for a visually impaired person may read the alt-text in place of an image, or a search engine may use the alt-text to better index a web page without analyzing the actual image content.

After extracting the alt-text, we perform language detection using CLD3 \citep{cld3} with three possible outputs: English, another language, or no detected language (i.e., all detections are below a confidence threshold \citep{schuhmann2021laion}).
Based on a manual inspection of a random sample, the ``no language'' set contains language-agnostic short form text such as the names of products and places.

We stored the resulting data in a PostgreSQL server for processing in the next stages of the pipeline.
We maintained about 500M image URLs in the server at all times.

\textbf{Downloading Image-Text Pairs.}
In order to maximize resource utilization, we downloaded the raw images from the parsed URLs with asynchronous requests using the Trio and Asks Python libraries.
To limit costs, we chose a small cloud node with 2 vCPUs, 1GB of RAM, and 10Mbps download bandwidth as a worker instance.
Such a worker can process 10,000 links in about 10 -- 15 minutes.
We utilized roughly 300 workers in parallel and batched the workload into chunks of 10,000 links taken from the aforementioned PostgreSQL server. 

\textbf{Post-Processing.}
After downloading the WAT files from Common Crawl, we removed data with less than 5 characters of text, less than 5 KB of image data, and potentially malicious, large, or redundant images.
To conclude the pipeline, we filtered image-text pairs based on their content.
Specifically, we computed cosine similarities between the image and text encodings with OpenAI's ViT-B/32 CLIP model.
For languages other than English, we utilized the multi-lingual CLIP ViT-B/32 from \citet{mclip}.
While OpenAI also released larger CLIP models later, these models were not available when we began to assemble LAION-5B.
For consistency, we therefore relied on ViT-B/32 CLIP models for the entire dataset.
We removed all English image-text pairs with cosine similarity below 0.28, and all other pairs with similarity below 0.26.
This step removed around 90\% of the original 50 billion images, leaving just short of 6 billion examples.
%

\subsection{Safety During Collection}
\label{subsec:safety_tagging}

%

Current automated filtering techniques are far from perfect: harmful images are likely to pass, and others are likely to be falsely removed.
We make a best effort to identify, document, and tag such content.
In the case of illegal content, we computed CLIP embeddings to filter out such samples.
Furthermore, these images and texts could amplify the social bias of machine learning models, especially ones trained with no or weak supervision \citep{steed2021image}.
It is important to note that the above mentioned classifiers are not perfect, especially keeping the complexity of these tasks and the diverse opinions of different cultures in mind. 
Therefore, we advocate using these tags responsibly, not relying on them to create a truly safe, ``production-ready'' subset after removing all potentially problematic samples. For a detailed discussion in this regard, we refer to Sec.~\ref{sec:ethical_discussion}.

To encourage research in fields such as dataset curation, we refrain from removing potentially offensive samples and tag them instead. The user can decide whether to include content depending on their task. To this end, we also encourage model developers to state, e.g., in their model card \citep{mitchell2019models} which subsets and tagged images are used. 

We apply Q16 \citep{schramowski2022can} and our own specialized pornographic and sexualized content classifier (here referred to as NSFW) to identify and document a broad range of inappropriate concepts displaying not only persons but also objects, symbols, and text, see \textit{cf.} \citep{schramowski2022can} and Appendix Sec.~\ref{sec:appendix_nsfwtagging} and Sec.~\ref{sec:appendix_q16tagging} 
for details. Both classifiers are based on CLIP embeddings.
Following our main intention of a publicly available dataset, these two approaches, as with all other implementations related to LAION 5B, are open-sourced.
%




We separate pornographic content and otherwise inappropriate content (e.g. harm, exploitation and degradation). Both can be dis- and enabled in the publicly available dataset exploration UI.\footnote{\href{https://github.com/rom1504/clip-retrieval}{https://knn5.laion.ai/}} With both together, the UI and the openly accessible code, we encourage users to explore and, subsequently, report further not yet detected content and thus contribute to the improvement of our and other existing approaches.

%% file: sections/composition.tex
\begin{figure}[!tb]
    \centering
    \subfloat{{\includegraphics[width=\linewidth]{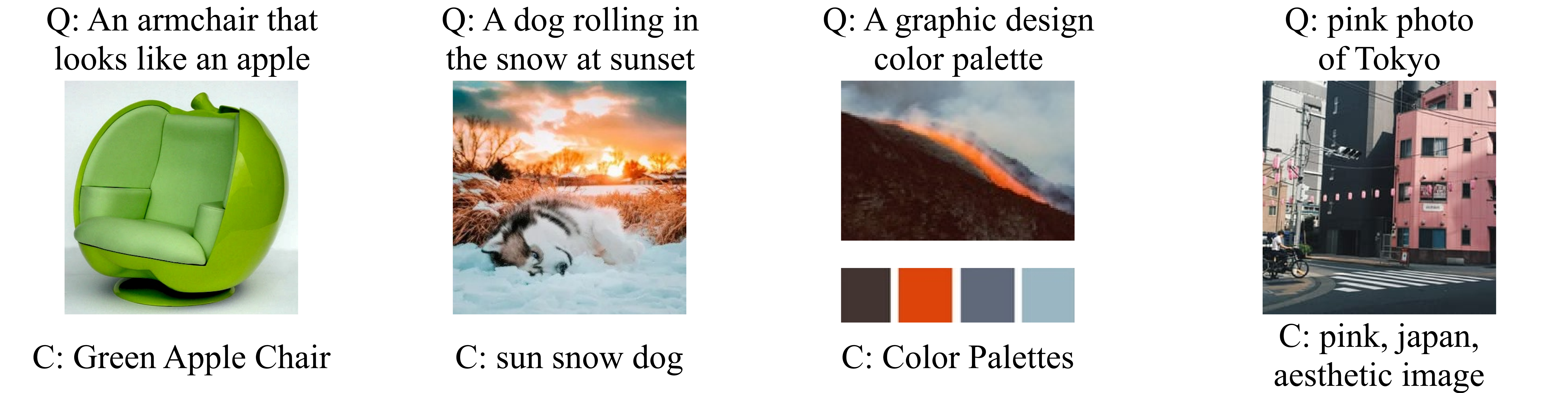}}}\\
    \caption{\textbf{LAION-5B examples.} Sample images from a nearest neighbor search in LAION-5B using CLIP embeddings. The image and caption (C) are the first results for the query (Q).}
    \label{sample}
\end{figure}

We release LAION-5B as the following three subsets:
\begin{itemize}
    \item 2.32 billion English image-text pairs. We refer to this subset as LAION-2B-en or LAION-2B if the language is clear from context.
    \item 2.26 billion image-text pairs from over 100 other languages.
    In the multilingual subset, the top-5 most frequent languages are Russian (10.6\%), French (7.4\%), German (6.6\%), Spanish (6.6\%), and Chinese (6.3\%).
    \item 1.27 billion samples where a language could not be clearly detected.
    Based on visually inspecting a random subset of these low-confidence language samples, the corresponding images often depict products or places.
    The captions contain language with clear semantics, but might also include noise such as keywords for search engine optimiziation or product tags.
\end{itemize}


We provide metadata files in the Apache Parquet format that consist of the following attributes for each image-text pair: 
\begin{itemize}[nosep]
    \item A 64-bit integer identifier
    \item The URL of the image. 
    \item The text string.
    \item Height and width of the image. 
    \item Cosine similarity between the text and image embeddings.
    \item The output from our NSFW and watermark detectors (one score between 0 and 1 each).
\end{itemize}
3\% of images were detected as NSFW, which can be filtered out by a user with the NSFW tag.

%% file: sections/experiments.tex

In this section, we showcase prior work using the LAION-400M \cite{schuhmann2021laion} and other subsets as well as our CLIP reproduction studies to give quantitative and qualitative evidence of the dataset's utility for training SOTA large scale language-vision models.

\subsection{Usage Examples}


\textbf{Subdataset Generation.} LAION-5B's scale enables novel dataset curation for computer vision related tasks. Recently, researchers have utilized both LAION-5B and a subset, LAION-400M, as a data source in vision related tasks such as facial representation learning \cite{LAION-400M-Faces} and invasive species mitigation \cite{lanternrd}. Within LAION, we have compiled from LAION-5B both LAION-High-Resolution\footnote{\href{https://huggingface.co/datasets/laion/laion-high-resolution}{https://huggingface.co/datasets/laion/laion-high-resolution}}, a 170M subset for superresolution models, and LAION-Aesthetic\footnote{\href{https://github.com/LAION-AI/laion-datasets/blob/main/laion-aesthetic.md}{https://github.com/LAION-AI/laion-datasets/blob/main/laion-aesthetic.md}}, a 120M subset of aesthetic images, as determined by a linear estimator on top of CLIP.

\textbf{CLIP Reproduction and Improvements.} \citet{LAION-400M-PyramidCLIP}, trained an enhanced CLIP architecture on the LAION-400M subset, outperforming OpenAI's CLIP on ImageNet zero-shot classification top-1 accuracy.
See Sec.~\ref{subsec:clip_exp} for our CLIP reproduction experiments using models of different scales. Training on a LAION-5B subset, \citet{LAION-400M-ImageCaptioning} developed BLIP to unify understanding and generation for vision-language tasks via a novel Vision-Language Pretraining (VLP) framework. It has been shown that BLIP matched or outperformed comparable models as per CIDEr, SPICE, and BLEU@4 metrics. \citet{LAION-400M-MAGMA} used a LAION subset for MAGMA, a model generating text “answers” for image-question pairs; MAGMA achieves state of the art results on OKVQA metrics and outperforming \textit{Frozen} \cite{tsimpoukelli2021multimodal}.



\textbf{Image Generation.} 
\citet{LAION-400M-LD} utilized a subset of LAION-5B in training latent diffusion models (LDM) that achieved state-of-the-art results on image inpainting and class-conditional image synthesis. The work was further extended into stable diffusion project that used subsets of LAION-5B (LAION-2B-en, laion-high-resolution and laion-aesthetics\footnote{See \url{https://github.com/CompVis/stable-diffusion} for more details}) for training a publicly available SOTA text-to-image generative model (see Appendix Sec. \ref{sec:appendix_experiments_stable_diffusion}). Furthermore, \citet{LAION-400M-Vector} used LAION-400M to train VQ diffusion text-to-image generation models, which have been shown to be more efficient, and are able to generate higher quality images.
Moreover, \citet{imagen} showed an improved architecture of a diffusion model that was trained on a subset of LAION-400M that outperforms OpenAI's recent DALLE-2 and achieves a new state-of-the-art COCO FID of 7.27.



\subsection{Experiments on CLIP Reproduction}
\label{subsec:clip_exp}

In an effort to reproduce the results of CLIP~\citep{radford2021learning}, and to validate the data collection pipeline we describe in Sec.~\ref{sec:collection_methodology}, we trained several models on LAION-400M~\citep{schuhmann2021laion} and a model on LAION-2B-en, datasets which are both subsets of LAION-5B. As training such models require large compute due to dataset and model sizes that are considered in the experiments, the usage of supercomputers and large compute clusters is necessary in order to train the models efficiently.

We used OpenCLIP~\citep{ilharco_gabriel_2021_5143773}, an open source software for training CLIP-like models. After adapting OpenCLIP for distributed training and execution on JUWELS Booster supercomputer \cite{JUWELSBooster2020}, we reproduced CLIP models of different size on the LAION-400M subset. We trained ViT-B/32, ViT-B/16, and ViT-L/14 following CLIP~\citep{radford2021learning}, and an additional model that we call ViT-B/16+, a slightly larger version of ViT-B/16.  
We followed the same hyper-parameter choices of the original CLIP models.
We used between 128 and 400 NVIDIA A100 GPUs to train the models. All trained models may be found in the OpenCLIP repository\footnote{\url{https://github.com/mlfoundations/open_clip}}.
For more information about hyper-parameters and training details, see Appendix Sec.~\ref{sec:appendix_experiments_training_details}.

\subsubsection{Zero-Shot Classification and Robustness Performance}

Following CLIP~\cite{radford2021learning} and subsequent works, we evaluate the models on zero-shot classification. For each downstream dataset, we use a set of pre-defined prompts for each class, which we collected from prior works~\cite{radford2021learning,zhai2021lit}.
We compute the embeddings of each class by averaging over the embedding of the prompts, computed each using the text encoder. For each image, and for each class, we compute the cosine similarity between their embeddings, and classify each image as the class that have the largest cosine similarity with the image embedding. We evaluate the models using top-1 accuracy.


In Tab.~\ref{table:zeroshot}, we show a comparison between models trained on LAION (400M, 2B) and original CLIP from~\cite{radford2021learning}. We follow~\cite{zhai2021lit} and evaluate robustness performance on ImageNet distribution shift datasets~\cite{pmlr-v97-recht19a, imagenetr, imagenetsketch, imageneta, objectnet}.
Additionally, we construct a benchmark we call VTAB+, a superset of VTAB~\cite{zhai2019large}, on which we compute the average top-1 accuracy over 35 tasks\footnote{\cite{zhai2019large} showed that different aggregation strategies have high rank correlation (Kendall score) with the simple top-1 average accuracy over datasets, thus we follow the same strategy. We also compute the ranks of each model on each task and average the ranks, and find that the ranking is similar to averaging top-1 accuracy.}. We can see that on ImageNet-1k (noted "INet" on the table), performance of LAION-400M models and original CLIP models (trained on a 400M private dataset) is matched well. On the four ImageNet distribution shift datasets, we observe some larger differences, notably on ObjNet (CLIP WIT is better) and INet-S (LAION is better), which allows us to conclude that in overall, CLIP models trained on LAION match in their robustness original CLIP.  With ViT-B/32 and ViT-L/14, training on the larger LAION-2B-en improves over LAION-400M model everywhere. 

\begin{figure}[!tb]
    \centering
    
    \subfloat{{\includegraphics[width=0.5\linewidth]{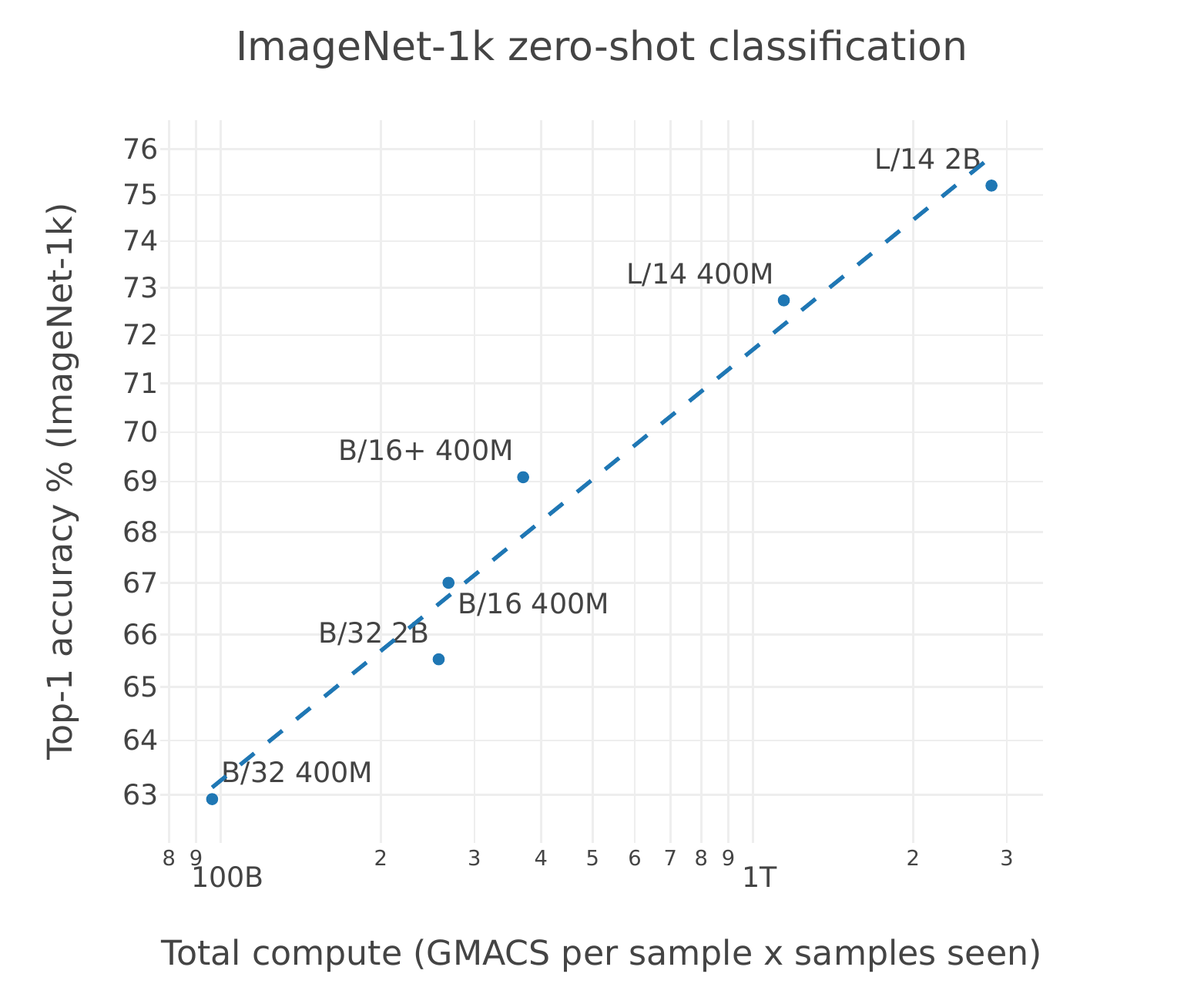}}}
    \subfloat{{\includegraphics[width=0.5\linewidth]{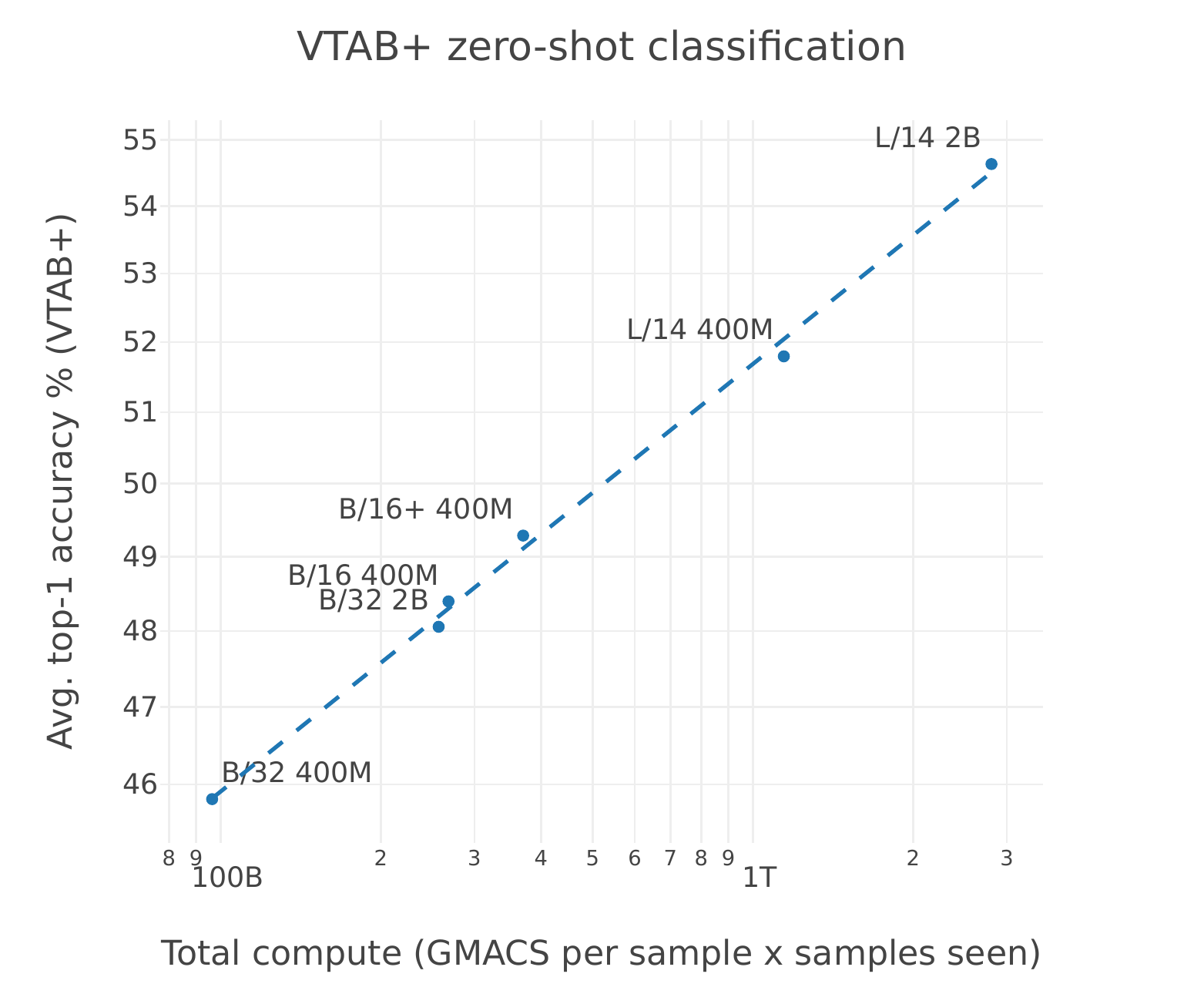}}}\\

    \caption{The relationship between total compute (giga multiply–accumulates (GMACS)) and zero-shot top-1 classification accuracy (\%) of models trained on LAION (400M, 2B-en). The dashed line in each figure is a linear fit in log-log space. Each point corresponds to a model trained on either the 400M or 2B-en LAION subsets. We show results on ImageNet-1k (left) and VTAB+ (right) where we average the accuracy over 35 tasks (see Appendix~\ref{sec:appendix_experiments_evaluation_details}  for details). Clear effect of model, data and compute training scale is evident on zero-shot performance that increases following scale power law.
    }
    \label{fig:compute_scale_clip}
\end{figure}

To obtain an idea about how the zero-shot performance improves with scale, we show the relationship between the total compute and accuracy on VTAB+ on models trained on LAION (400M, 2B-en). In Figure \ref{fig:compute_scale_clip}, we  see that accuracy on VTAB+ improves with compute (log-log plot). It would be interesting to study in future work if the relationship between compute and accuracy keeps showing the same trend or whether we start to see saturation, like it was observed in~\cite{zhai2021scaling}. Here, we can report that increasing either model or data scale for CLIP pre-training results in improvement of zero-shot classification performance on various downstream transfer targets. For a full overview of zero-shot classification and retrieval results, view Sec.~\ref{sec:appendix_experiments_evaluation_details} of the Appendix.


To show that larger dataset scale matters for the performance of pre-trained models, we perform additional experiments using  ViT-B/32 and ViT-L/14 on different LAION-5B and LAION-400M subsets, while varying the amount of training compute (samples seen). Our findings confirm that the effect of dataset scale is significant, given sufficient compute for training. For instance, for the same amount of compute (34B images seen), training ViT-L/14 on LAION-2B-en (75.4\%) outperforms LAION-400M (73.9\%) on ImageNet-1k zero-shot classification. Same effect is observed for smaller ViT-B/32 model. For more detailed results, see Fig.~\ref{fig:data_scale} and Tab.~\ref{table:data_scale} in the Appendix.  



\begin{table}[t]
\centering
\begin{small}
\rowcolors{2}{light-light-gray}{white}
\begin{tabular}{@{}llllllll@{}}
\toprule
\bf{Model} & \bf{Pre-training}  & \hspace{0.18cm}\rotatebox{90}{INet}&\hspace{0.18cm}\rotatebox{90}{INet-v2}&\hspace{0.18cm}\rotatebox{90}{INet-R}&\hspace{0.18cm}\rotatebox{90}{INet-S}&\hspace{0.18cm}\rotatebox{90}{ObjNet}&\hspace{0.18cm}\rotatebox{90}{VTAB+}\\ \midrule
 \multirow{1}{*}{B/32}& CLIP WIT & 63.3&56.0&69.4&42.3&44.2&45.4\\
 & LAION-400M&${62.9}^{\tiny\color{red}\textbf{-0.4}}$&${55.1}^{\tiny\color{red}\textbf{-0.9}}$&${73.4}^{\tiny\color{ForestGreen}\textbf{+4.0}}$&${49.4}^{\tiny\color{ForestGreen}\textbf{+7.1}}$&${43.9}^{\tiny\color{red}\textbf{-0.3}}$&${45.6}^{\tiny\color{ForestGreen}\textbf{+0.2}}$\\
 & LAION-2B-en&${65.7}^{\tiny\color{ForestGreen}\textbf{+2.4}}$&${57.4}^{\tiny\color{ForestGreen}\textbf{+1.4}}$&${75.9}^{\tiny\color{ForestGreen}\textbf{+6.5}}$&${52.9}^{\tiny\color{ForestGreen}\textbf{+10.6}}$&${48.7}^{\tiny\color{ForestGreen}\textbf{+4.5}}$&${47.9}^{\tiny\color{ForestGreen}\textbf{+2.5}}$\\\midrule
 \multirow{1}{*}{B/16} & CLIP WIT & 68.3&61.9&77.7&48.2&55.3&47.5\\
 & LAION-400M&${67.0}^{\tiny\color{red}\textbf{-1.3}}$&${59.6}^{\tiny\color{red}\textbf{-2.3}}$&${77.9}^{\tiny\color{ForestGreen}\textbf{+0.2}}$&${52.4}^{\tiny\color{ForestGreen}\textbf{+4.2}}$&${51.5}^{\tiny\color{red}\textbf{-3.8}}$&${48.3}^{\tiny\color{ForestGreen}\textbf{+0.8}}$\\\midrule
\multirow{1}{*}{B/16+} & LAION-400M & 69.2&61.5&80.5&54.4&53.9&49.2\\\midrule
 \multirow{1}{*}{L/14}& CLIP WIT & 75.6&69.8&87.9&59.6&69.0&55.7\\
 & LAION-400M&${72.8}^{\tiny\color{red}\textbf{-2.8}}$&${65.4}^{\tiny\color{red}\textbf{-4.4}}$&${84.7}^{\tiny\color{red}\textbf{-3.2}}$&59.6&${59.9}^{\tiny\color{red}\textbf{-9.1}}$&${51.8}^{\tiny\color{red}\textbf{-3.9}}$\\
 & LAION-2B-en&${\color{black}{75.2}}^{\tiny\color{red}\textbf{-0.3}}$&${\color{black}{67.7}}^{\tiny\color{red}\textbf{-2.0}}$&${\color{black}{87.4}}^{\tiny\color{red}\textbf{-0.5}}$&${\color{black}{63.3}}^{\tiny\color{ForestGreen}\textbf{+3.7}}$&${\color{black}{65.5}}^{\tiny\color{red}\textbf{-3.6}}$&${\color{black}{54.6}}^{\tiny\color{red}\textbf{-1.2}}$\\
 \midrule
\end{tabular}
\caption{Comparison between CLIP models trained on LAION (400M, 2B) and the original CLIP models \cite{radford2021learning} trained on OpenAI's WebImageText (WIT) dataset. We show zero-shot top-1 classification accuracy (\%) on various datasets including ImageNet, four ImageNet  distribution shift datasets, and a benchmark we call VTAB+, where we average performance over 35 tasks. See Appendix~\ref{sec:appendix_experiments_evaluation_details} 
for more details about the datasets used for evaluation and the results.}
\label{table:zeroshot}
\end{small}
\end{table}

\subsection{Experiments with Generative Models}
\label{subsec:laionide_exp}
To validate LAION-5B as a dataset for training strong text-to-image generation models, we fine-tuned OpenAI's GLIDE \cite{GLIDE} on LAION-5B data. The obtained results comparing generated samples from original OpenAI GLIDE and from our reproduction (LAIONIDE) are compiled into an interactive web demo\footnote{\href{https://wandb.ai/afiaka87/glide_compare/reports/laionide-v3-benchmark--VmlldzoxNTg3MTkz}{https://wandb.ai/afiaka87/glide\_compare/reports/laionide-v3-benchmark--VmlldzoxNTg3MTkz}}. See Appendix Sec \ref{sec:appendix_experiments_generative} for more technical details on experiments with GLIDE (\ref{sec:appendix_experiments_GLIDE}) and Stable Diffusion (\ref{sec:appendix_experiments_stable_diffusion}). 



%% file: sections/strengths_and_weaknesses.tex
The large scale of current image-text datasets makes it infeasible to thoroughly investigate all aspects of a dataset in a single publication.
Hence we now outline some potential technical limitations specifically affecting LAION-5B.
These potential limitations are starting points for future work on analyzing and improving image-text datasets.

\textbf{Data Overlap.} Our experiments in Section \ref{subsec:clip_exp} show that models trained on LAION-5B achieve good performance on a variety of downstream tasks.
However, the LAION-5B training set may overlap with some of the downstream test sets if these test sets are also included in Common Crawl.
If overlap is present, it may lead to incorrectly large test set accuracies that overstate the true generalization capabilities of models trained on LAION-5B.

Overall, we do not consider potential test set overlap to be a serious threat for the validity of results obtained with LAION-5B.
OpenAI encountered the same question in the context of their pre-training dataset for CLIP and found only few examples of substantial performance difference due to data overlap on downstream target datasets \citep{radford2021learning}.
Some datasets such as ObjectNet~\cite{objectnet} are likely not contained in Common Crawl because ObjectNet was not assembled from web images.
Instead, the authors of ObjectNet tasked MTurk workers to take new pictures in their own homes.
Nevertheless, measuring the degree of overlap between LAION-5B and popular computer vision benchmarks is an important question for future work, which will include further de-duplication efforts.

\textbf{Other text sources.}
\citet{birhane2021multimodal} described the shortcomings of alt-text and noted that alt-text is not necessarily a good description of the corresponding image.
For instance, the alt-text may be search engine optimization (SEO) spam, an incoherent list of keywords, or overly corrupted otherwise. In such cases, the language in the text annotations may become less informative or entirely useless for training.
For ImageNet zero-shot classification,  BASIC \citep{basic} has demonstrated strong results when turning 5 billion of the 6.6 billion captions into the form of \texttt{CLASS\_1} \texttt{and} \texttt{CLASS\_2} \texttt{and}
 \texttt{...} \texttt{and} \texttt{CLASS\_K}, by using an internal multi-label classification dataset (JFT-3B). Thus, image captions formed by just concatenating class names may also serve as meaningful alternative of otherwise corrupted text. Such a finding adds a possibility of employing generated together with existing natural language captions for training contrastive image-language models with strong zero-shot performance.
 

\textbf{Filtering with CLIP.}
CLIP allows the curation and collection of this dataset to be low-cost and scalable. Such an automated process reduces dramatically necessity for the human control which would be otherwise intractable for such large scale collection. However, through curating with CLIP, we also incur its flaws and model biases. For additional discussion of CLIP filtering related to safety and ethics, see Appendix Sec. \ref{sec:appendix_clip_filtering}.

Filtering by a small scale CLIP ViT-B/32 may leave more image-text pairs with weak or no semantic connection in the dataset while also accidentally removing some high quality image-text pairs than filtering with stronger, larger scale models that were not available in the time of our experiments. The larger CLIP ViT-L/14 model may create a less noisy version of LAION datasets than what was possible with smaller scale CLIP ViT-B/32. We hypothesize that filtering Common Crawl with a CLIP ViT-L model will further increase the quality of our dataset. It is subject to our future work to create a CLIP ViT L/14 filtered version of LAION-400M and LAION-5B to test how this affects model training and downstream transfer performance. 




%% file: sections/ethical_social.tex




Recent developments in large-scale models, such as GPT-3 \citep{GPT-3}, CLIP \citep{CLIP}, ALIGN \citep{ALIGN}, GLIDE \citep{GLIDE}, and DALLE-2 \citep{DALLE-2} have potential for far-reaching impact on society, both positive and negative, when deployed in applications such as image classification and generation, recommendation systems, or search engines.
Besides model parameter scaling, the advances made so far also rely on the underlying large-scale datasets.
Recent research \citep{bender2021Stochastic, birhane2021Large} described many potential negative societal implications that may arise due to careless use of vision-language models, e.g., the models perform worse for certain groups of users or reproduce discriminatory behavior. 

Unfortunately, only a minority of these models are publicly released, most of them are only accessible by an ``input to output'' interface. Importantly, the underlying large-scale datasets are also not often publicly available. 
While open-source efforts exist to re-implement model architectures and training, the closed nature of large-scale datasets used for model training makes any proper systematic investigation of model training and model behavior very hard or even impossible. Studying full training, comparison of different model architectures and progress in large-scale multi-modal learning becomes restricted to those institutions that were able to obtain their closed large-scale datasets. It also results in safety issues of creating and using such models, as broad research community does not get to test both model and the dataset used for its training for causes underlying undesired behaviours. 

LAION-5B as an open large-scale dataset provides here not only a chance to make progress in careful studies of the trained models' capabilities and replication but also to investigate how uncurated large-scale datasets impact various model biases and under which circumstances their usage may result in undesired safety issues. Such research can help to design automated ways to curate and create datasets from uncurated ones that alleviate the bias and safety issues. To this end, LAION also created a number of tools to aid researchers and other users in large-scale data handling and exploration. One such a tool uses pre-computed image embeddings to enable search of images guided either by text or image input via an easily and publically accessible web interface (CLIP retrieval tool\footnote{\href{https://knn5.laion.ai}{https://knn5.laion.ai}}, see Appendix Sec. \ref{sec:appendix_ui_search}). LAION made also source code for the tool and routines necessary to build an own version of it publicly available\footnote{\href{https://github.com/rom1504/clip-retrieval}{https://github.com/rom1504/clip-retrieval}} (see Appendix Sec \ref{sec:appendix_preparation_curation}, \ref{sec:appendix_dist_inference}, \ref{sec:appendix_dist_indexing} for more details).

After the release of LAION-400M, several groups (e.g., \citep{birhane2021multimodal}) already used such tools and investigated potential problems arising from an unfiltered dataset. Motivated by these findings, with LAION-5B, we introduced an improved inappropriate content tagging (\textit{cf.} Sec.~\ref{subsec:safety_tagging}) 
as well as a watermark filter, which can improve the safety and quality of the text-to-image models trained on the dataset.

Such development indicates that this dataset acts as a starting point, and is not the final endpoint, for creating further improved datasets to train models for various tasks. In our opinion, this process is not supposed to be a non-transparent closed-door avenue. It should be approached by broad research community, resulting in open and transparent datasets and procedures for model training. Towards meeting this challenge, the large-scale public image-text dataset of over 5.8 billion pairs and further annotations introduced here provides diversity that can be a starting point for ensuring balance and for selecting safe, curated subsets for corresponding target applications. We encourage everybody to participate in this exciting and important future journey.

In the current form, we consider this dataset a research artefact and strongly advocate \textbf{academic use-only} and advise careful investigation of downstream model biases (Appendix Sec. \ref{sec:appendix_clip_filtering}). Additionally, we encourage users to use the described tools and to transparently explore and, subsequently, report further not yet detected content and model behaviour to our dataset repository\footnote{\href{https://github.com/laion-ai/laion5b-bias}{https://github.com/laion-ai/laion5b-bias}}, and help to further advance existing approaches for data curation using the real-world large dataset introduced here.

\textbf{Privacy.} We comment on privacy issues arising from Common Crawl as source of links in LAION-5B and measures undertaken to handle those in the Appendix Sec. \ref{sec:appendix_privacy}


%% file: sections/conclusion.tex

By releasing LAION-5B, a larger updated version of an openly available dataset that contains over 5 billion image-text pairs, we have further pushed the scale of open datasets for training and studying state-of-the-art language-vision models. This scale gives strong increases to zero-shot transfer and robustness.

To validate the utility of LAION-5B, we demonstrated that a subset of our dataset can be used to train SOTA CLIP models of various scale that match the strong zero-shot and robustness performance of the original models trained on closed curated data, or to fine-tune generative models like GLIDE, producing samples of good quality. The dataset thus provides opportunities in multi-language large-scale training and research of language-vision models, that were previously restricted to those having access to proprietary large datasets, to the broader research community. Finally, thanks to its large scale, even a rather strict subset filtering (driven by various criterion like NSFW, watermark presence, resolution) provides high-quality datasets that are still large enough to provide sufficient scale for the training or fine-tuning of strong specialized language-vision models. 

%% file: sections/acknowledgments.tex

We thank Phil Wang, the creator of the DALLE-pytorch github repository\footnote{\url{https://github.com/lucidrains/DALLE-pytorch}}, who inspired us and helped creating our open community. We also want to thank Aran Komatsuzaki, Andreas Köpf, Bokai Yu, John David Pressman, Natalie Parde, Gabriel Ilharco, Fredde Frallan (see also Appendix) and all the members of the LAION discord server\footnote{\url{https://discord.gg/xBPBXfcFHd}} for helping crawling image-text-pairs and run inference on their private computers. We want to thank Hugging Face and Stability AI for their continuous financial support and providing hosting space for open datasets and models. We would also like to thank openAI for making their pre-trained CLIP models publicly available, which allowed us to filter the LAION datasets. We would like to express gratitude to all the people who are working on making code, models and data publicly available, advancing community based research and making research more reproducible.

The authors gratefully acknowledge the Gauss Centre for Supercomputing e.V. \footnote{\url{https://gauss-centre.eu}} for funding this work by providing computing time through the John von Neumann Institute for Computing (NIC) on the GCS Supercomputer JUWELS Booster \citep{JUWELSBooster2020} at Jülich Supercomputing Centre (JSC). We also acknowledge storage resources on JUST \citep{graf2021just} granted and operated by JSC.
Patrick Schramowski acknowledges the support by the Hessian Ministry of Higher Education, Research, Science and the Arts (HMWK) cluster project ``The Third Wave of AI''.

%% file: sections/references.tex

\bibliographystyle{plainnat}
\bibliography{references}

%% file: sections/datasheet.tex

\subsection{Motivation}

\begin{enumerate}[label=Q\arabic*]

\item \textbf{For what purpose was the dataset created?} Was there a specific task in mind? Was there a specific gap that needed to be filled? Please provide a description.

\begin{itemize}
\item LAION-5B was created as an open solution to training very large multimodal models such as CLIP or DALL-E. Before the curation of this dataset, the closest in size was YFCC with 100 million image/videos and associated metadata. OpenAI previously used a 15 million sample subset to train a publicly comparable CLIP model, but that pales in comparison to the private 400 million sample dataset they used to train the high-performant CLIP models. At the time of writing this, the ImageNet-1k zero-shot top-1 state-of-the-art, Google’s BASIC, used a dataset of 6.6 billion image-text pairs. With the release of LAION-5B, researchers no longer have to be part of a few selected institutions to study these problems.
\end{itemize}

\item \textbf{Who created the dataset (e.g., which team, research group) and on behalf of which entity (e.g., company, institution, organization)?}

\begin{itemize}
\item This dataset is presented by LAION (Large-scale Artificial Intelligence Open Network), a non-profit research organization aiming to democratize access to large-scale open datasets and powerful machine learning models through the research and development of open-source resources. The communication and organization of this project took place on the open LAION discord server \footnote{https://discord.gg/xBPBXfcFHd}.
\end{itemize}

\item \textbf{Who funded the creation of the dataset?} If there is an associated grant, please provide the name of the grantor and the grant name and number.

\begin{itemize}
\item This work was sponsored by Hugging Face and Stability AI.
\end{itemize}

\item \textbf{Any other comments?}

\begin{itemize}
\item No.
\end{itemize}

\subsection{Composition}

\item \textbf{What do the instances that comprise the dataset represent (e.g., documents, photos, people, countries)?} \textit{Are there multiple types of instances (e.g., movies, users, and ratings; people and interactions between them; nodes and edges)? Please provide a description.}

\begin{itemize}
\item We provide 5.8 billion image-text pairs. Each pair consists of the following: an image file url; text caption; width; height; the caption’s language; cosine similarity (CLIP ViT/B-32 for English and MCLIP for multiple and unknown languages); the probability of the image containing a watermark; the probability of a sample being NSFW. We made our models openly available on the LAION github page (\url{https://github.com/LAION-AI/LAION-5B-WatermarkDetection}, \url{https://github.com/LAION-AI/CLIP-based-NSFW-Detector}).
\end{itemize}

\item \textbf{How many instances are there in total (of each type, if appropriate)?}

\begin{itemize}
\item LAION-5B contains 2.3 billion English samples, 2.2 billion multilingual samples, and 1.2 billion unknown language samples. A further overview of the statistics may be seen in the announcement \href{https://laion.ai/laion-5b-a-new-era-of-open-large-scale-multi-modal-datasets/}{blog post} .
\end{itemize}

\item \textbf{Does the dataset contain all possible instances or is it a sample (not necessarily random) of instances from a larger set?} \textit{If the dataset is a sample, then what is the larger set? Is the sample representative of the larger set (e.g., geographic coverage)? If so, please describe how this representativeness was validated/verified. If it is not representative of the larger set, please describe why not (e.g., to cover a more diverse range of instances, because instances were withheld or unavailable).}

\begin{itemize}
\item Common Crawl is a public repository of crawled web pages. From this collection of web pages we filter the images and alt-text to derive LAION-5B. Of the existing 50+ billion images available in common crawl. We provide image url and alt-text pairings of only 5.8 billion images.
\end{itemize}

\item \textbf{What data does each instance consist of?} \textit{“Raw” data (e.g., unprocessed text or images) or features? In either case, please provide a description.}

\begin{itemize}
\item We provide raw urls and their associated alt-text.
\end{itemize}

\item \textbf{Is there a label or target associated with each instance?} \textit{If so, please provide a description.}

\begin{itemize}
\item There is no hard class label, but researchers will often formulate a mapping of the text to image or vice-versa.
\end{itemize}

\item \textbf{Is any information missing from individual instances?} \textit{If so, please provide a description, explaining why this information is missing (e.g., because it was unavailable). This does not include intentionally removed information, but might include, e.g., redacted text.}

\begin{itemize}
\item No.
\end{itemize}

\item \textbf{Are relationships between individual instances made explicit (e.g., users' movie ratings, social network links)?} \textit{If so, please describe how these relationships are made explicit.}

\begin{itemize}
\item No.
\end{itemize}

\item \textbf{Are there recommended data splits (e.g., training, development/validation, testing)?} \textit{If so, please provide a description of these splits, explaining the rationale behind them.}

\begin{itemize}
\item No.
\end{itemize}

\item \textbf{Are there any errors, sources of noise, or redundancies in the dataset?} \textit{If so, please provide a description.}

\begin{itemize}
\item There exist near duplicate images which makes possible a many to one embedding in certain scenarios. CLIP embeddings may be used to remove more or less of them.
\end{itemize}

\item \textbf{Is the dataset self-contained, or does it link to or otherwise rely on external resources (e.g., websites, tweets, other datasets)?} \textit{If it links to or relies on external resources, a) are there guarantees that they will exist, and remain constant, over time; b) are there official archival versions of the complete dataset (i.e., including the external resources as they existed at the time the dataset was created); c) are there any restrictions (e.g., licenses, fees) associated with any of the external resources that might apply to a future user? Please provide descriptions of all external resources and any restrictions associated with them, as well as links or other access points, as appropriate.}

\begin{itemize}
\item This dataset is reliant on links to the World Wide Web. As such, we are unable to offer any guarantees of the existence of these samples. Due to the size we will also not be able to offer archives of the current state either. In order to rapidly and efficiently download images from URLs, we provide \href{https://github.com/rom1504/img2dataset}{img2dataset}. Depending on bandwidth, it’s feasible to download the entire LAION-5B dataset in 7 days using 10 nodes.
\end{itemize}

\item \textbf{Does the dataset contain data that might be considered confidential (e.g., data that is protected by legal privilege or by doctor–patient confidentiality, data that includes the content of individuals’ non-public communications)?} \textit{If so, please provide a description.}

\begin{itemize}
\item This dataset was collected using openly available parts of the internet with the assumption that any data found was intended to be shared freely. However, it is possible that the parties crawled by Common Crawl may have publicly hosted confidential data.
\end{itemize}

\item \textbf{Does the dataset contain data that, if viewed directly, might be offensive, insulting, threatening, or might otherwise cause anxiety?} \textit{If so, please describe why.}

\begin{itemize}
\item Since the dataset is scraped from Common Crawl, it is known to have instances of sexually explicit, racist, abusive or other discomforting or disturbing content. We choose to include these samples for the usage of safety researchers and further dataset curation surrounding these sensitive topics.   
\item To address the existence of distressing content, we provide safety tags. Details on tagging potentially inappropriate content can be found in Sec.~\ref{subsec:safety_tagging} in the main text and Appendix Sec.~\ref{sec:appendix_nsfwtagging} and Sec.~\ref{sec:appendix_q16tagging}.
During down-stream training tasks, users may check the sample’s boolean flags to determine whether or not the sample should be used. 
However, as we described in the main text, it is important to note that the safety tags are not perfect, especially keeping the complexity of these tasks and the diverse opinions of different cultures in mind. Therefore, we advocate using these tags responsibly, not relying on them to create a truly safe, ``production-ready'' subset after removing all potentially problematic samples.
\end{itemize}

\item \textbf{Does the dataset relate to people?} \textit{If not, you may skip the remaining questions in this section.}

\begin{itemize}
\item People may be present in the images or textual descriptions, but people are not the sole focus of the dataset.
\end{itemize}

\item \textbf{Does the dataset identify any subpopulations (e.g., by age, gender)?}

\begin{itemize}
\item We do not provide any markers of subpopulation as attributes of the image-text pairs, but it may be possible to deduce this in some cases from the image and language pairing.
\end{itemize}

\item \textbf{Is it possible to identify individuals (i.e., one or more natural persons), either directly or indirectly (i.e., in combination with other data) from the dataset?} \textit{If so, please describe how.}

\begin{itemize}
\item Yes it may be possible to identify people using face recognition. We do not provide any such means nor make attempts, but institutions owning large amounts of face identifiers may identify specific people in the dataset. Similarly, people may be identified through the associated text.
\end{itemize}

\item \textbf{Does the dataset contain data that might be considered sensitive in any way (e.g., data that reveals racial or ethnic origins, sexual orientations, religious beliefs, political opinions or union memberships, or locations; financial or health data; biometric or genetic data; forms of government identification, such as social security numbers; criminal history)?} \textit{If so, please provide a description.}

\begin{itemize}
\item Yes the dataset contains sensitive content. Although, the dataset wasn’t created with the intention of obtaining samples fitting this criteria, it is possible that individuals might have hosted such items on a website that had been crawled by Common Crawl.
\end{itemize}

\item \textbf{Any other comments?}

\begin{itemize}
\item We caution discretion on behalf of the user and call for responsible usage of the dataset for research purposes \textbf{only}. 
\end{itemize}

\subsection{Collection Process}

\item \textbf{How was the data associated with each instance acquired?} \textit{Was the data directly observable (e.g., raw text, movie ratings), reported by subjects (e.g., survey responses), or indirectly inferred/derived from other data (e.g., part-of-speech tags, model-based guesses for age or language)? If data was reported by subjects or indirectly inferred/derived from other data, was the data validated/verified? If so, please describe how.}

\begin{itemize}
\item From the aforementioned Common Crawl, we filter images and their associated alt-text. Inclusion is determined by cosine similarity of the alt-text and the image as determined by OpenAI’s CLIP ViT-B/32 for english samples and MCLIP for all other samples. We include English samples with a cosine similarity score above 0.28, and we select all multilingual and unknown language samples with a 0.26 cosine similarity score or greater.
\end{itemize}

\item \textbf{What mechanisms or procedures were used to collect the data (e.g., hardware apparatus or sensor, manual human curation, software program, software API)?} \textit{How were these mechanisms or procedures validated?}

\begin{itemize}
\item We ran a preprocessing script in python, over hundred of small CPU nodes, and few GPU nodes. They were validated by manual inspection of the results and post processing on them: computation of statistics on the width, height, size of captions, clip embeddings and indices
\end{itemize}

\item \textbf{If the dataset is a sample from a larger set, what was the sampling strategy (e.g., deterministic, probabilistic with specific sampling probabilities)?}

\begin{itemize}
\item The dataset was obtained by openAI CLIP ViT B/32 filtering of Common Crawl links using cosine similarity of the image and its text the links were referring to.
\end{itemize}

\item \textbf{Who was involved in the data collection process (e.g., students, crowdworkers, contractors) and how were they compensated (e.g., how much were crowdworkers paid)?}

\begin{itemize}
\item No crowdworkers were used in the curation of the dataset. Open-source researchers and developers enabled its creation for no payment.
\end{itemize}

\item \textbf{Over what timeframe was the data collected? Does this timeframe match the creation timeframe of the data associated with the instances (e.g., recent crawl of old news articles)?} \textit{If not, please describe the timeframe in which the data associated with the instances was created.}

\begin{itemize}
\item The data was filtered from September 2021 to January 2022, but those who created the sites might have included content from before then. It is impossible to know for certain how far back the data stretches.
\end{itemize}

\item \textbf{Were any ethical review processes conducted (e.g., by an institutional review board)?} \textit{If so, please provide a description of these review processes, including the outcomes, as well as a link or other access point to any supporting documentation.}

\begin{itemize}
\item We corresponded with the University of Washington's Human Subject Division, and as we do not intervene with the people depicted in the data as well as the data being public, they stated that the work did not require IRB review. Furthermore, the NeurIPS ethics review determined that the work has no serious ethical issues.  
\end{itemize}

\item \textbf{Does the dataset relate to people?} \textit{If not, you may skip the remaining questions in this section.}

\begin{itemize}
\item People may appear in the images and descriptions, although they are not the exclusive focus of the dataset. 
\end{itemize}

\item \textbf{Did you collect the data from the individuals in question directly, or obtain it via third parties or other sources (e.g., websites)?}

\begin{itemize}
\item We retrieve the data from Common Crawl which contains almost all websites.
\end{itemize}

\item \textbf{Were the individuals in question notified about the data collection?} \textit{If so, please describe (or show with screenshots or other information) how notice was provided, and provide a link or other access point to, or otherwise reproduce, the exact language of the notification itself.}

\begin{itemize}
\item Individuals were not notified about the data collection.
\end{itemize}

\item \textbf{Did the individuals in question consent to the collection and use of their data?} \textit{If so, please describe (or show with screenshots or other information) how consent was requested and provided, and provide a link or other access point to, or otherwise reproduce, the exact language to which the individuals consented.}

\begin{itemize}
\item We follow Common Crawl’s practice of crawling the web and follow each site’s robots.txt file, thus users consent to their sites being crawled. However, those depicted in the photograph might not have given their consent to its upload.
\end{itemize}

\item \textbf{If consent was obtained, were the consenting individuals provided with a mechanism to revoke their consent in the future or for certain uses?} \textit{If so, please provide a description, as well as a link or other access point to the mechanism (if appropriate).}

\begin{itemize}
\item Users have a possibility to check for the presence of the links in our dataset leading to their data on public internet by using the search tool provided by LAION, accessible at \href{https://knn5.laion.ai}{https://knn5.laion.ai}. If users wish to revoke their consent after finding sensitive data, they can contact the hosting party and request to delete the content from the underlying website – it will be automatically removed from LAION-5B since we distributed image-text pairs as URLs. Moreover, we provide a contact email \texttt{contact@laion.ai} and contact form \href{https://laion.ai/dataset-requests/}{https://laion.ai/dataset-requests/} to request removal of the links from the dataset. The actual content behind the links is out of our reach and will in that case remain accessible on the public internet for other crawlers.
\end{itemize}

\item \textbf{Has an analysis of the potential impact of the dataset and its use on data subjects (e.g., a data protection impact analysis) been conducted?} \textit{If so, please provide a description of this analysis, including the outcomes, as well as a link or other access point to any supporting documentation.}

\begin{itemize}
\item \href{https://arxiv.org/abs/2110.01963}{Birhane, Prabhu, and Kahembwe} opened the discussion on the limitations and imminent biases that come with the creation of a weakly-curated dataset using CLIP. CLIP and its usage of cosine similarity offers a useful but imperfect heuristic for dataset inclusion that inherits various biases contained in the image-text pairs crawled from the web. In addition, the biases already existent within CLIP and the World Wide Web may become amplified when distilling original raw data and forming a filtered dataset. Using a model trained on this dataset without any further curation in production has the potential to reinforce harmful simplistic stereotypes against already marginalized communities.
\item However, the authors also note that this dataset posits currently the only openly available solution for studying multimodal models of this scale, examining their potential benefits and harms. Combining the aforementioned limitations and opportunities that this dataset provides, we agree with the authors and authorize the dataset for purely academic endeavors and strongly advice against any usage in end products.

\end{itemize}

\item \textbf{Any other comments?}

\begin{itemize}
\item No.
\end{itemize}

\subsection{Preprocessing, Cleaning, and/or Labeling}

\item \textbf{Was any preprocessing/cleaning/labeling of the data done (e.g., discretization or bucketing, tokenization, part-of-speech tagging, SIFT feature extraction, removal of instances, processing of missing values)?} \textit{If so, please provide a description. If not, you may skip the remainder of the questions in this section.}

\begin{itemize}
\item No preprocessing or labelling is done. Certain images were removed on the basis of safety, and others are tagged in the presence of NSFW content or a watermark.
\end{itemize}

\item \textbf{Was the “raw” data saved in addition to the preprocessed/cleaned/labeled data (e.g., to support unanticipated future uses)?} \textit{If so, please provide a link or other access point to the “raw” data.}

\begin{itemize}
\item We do not save the raw data.
\end{itemize}

\item \textbf{Is the software used to preprocess/clean/label the instances available?} \textit{If so, please provide a link or other access point.}

\begin{itemize}
\item To preprocess the data we used:
\begin{itemize}
\item \url{https://github.com/rvencu/crawlingathome-gpu-hcloud} process common crawl into a laion5B-like dataset
\item \url{http://github.com/rom1504/img2dataset} A tool to easily turn large sets of image urls to an image dataset. Can download, resize and package 100M urls in 20h on one machine.
\item \url{https://github.com/rom1504/clip-retrieval} a tool to easily compute clip embeddings and build a clip retrieval system with them
\end{itemize}
\item For individuals to preprocess the data for training, we provide:
\begin{itemize}
    \item \url{https://github.com/rom1504/laion-prepro}
\end{itemize}
\end{itemize}

\item \textbf{Any other comments?}

\begin{itemize}
\item No.
\end{itemize}

\subsection{Uses}

\item \textbf{Has the dataset been used for any tasks already?} \textit{If so, please provide a description.}

\begin{itemize}
\item LAION-5B (and the associated LAION-400M) has been used on a number of tasks such as CLIP Reproduction, BLIP Training, Glide Training, Cloob Training, and sub-dataset generation. For example, \href{https://arxiv.org/abs/2111.14822.pdf}{Gu et al.} used LAION-400M to train VQ diffusion text-to-image generation models. Additionally, \href{https://arxiv.org/abs/2112.10752.pdf}{Rombach et al.} applied a subset of LAION-400M in training Latent Diffusion Models that achieved state-of-the-art results on image inpainting and class-conditional image synthesis. The team behind \href{https://github.com/mlfoundations/open\_clip}{open\_CLIP} demonstrated the capabilities of the 400M subset for CLIP reproduction, achieving performance on par with that of OpenAI. On the matter of subset generation and CLIP reproduction, \href{https://arxiv.org/abs/2112.03109.pdf}{Zheng et al.} utilized LAION for facial representation learning. It should be noted that this example demonstrates the potential for users to misuse this dataset for the purpose of identification. \href{https://arxiv.org/abs/2201.12086}{Li et al.} applied a subset of LAION for the purpose of image-captioning. Finally, \href{https://arxiv.org/pdf/2112.05253.pdf}{Eichenberg et al.} used a LAION subset for MAGMA, a model generating text “answers'' for image-question pairs. 
\end{itemize}

\item \textbf{Is there a repository that links to any or all papers or systems that use the dataset?} \textit{If so, please provide a link or other access point.}

\begin{itemize}
\item Yes, scientific publications and systems that use LAION datasets can be found on the \href{https://github.com/LAION-AI/dataset-usage }{LAION github page}.
\end{itemize}

\item \textbf{What (other) tasks could the dataset be used for?}

\begin{itemize}
\item We encourage future researchers to curate LAION-5B for several tasks. Particularly, we see applications of the dataset in image and text representation learning, image to text generation, image captioning, and other common multimodal tasks. Due to the breadth of the data, it also offers a unique opportunity for safety and low resource language researchers. We hope for LAION-5B to serve under-represented projects as well.
\end{itemize}

\item \textbf{Is there anything about the composition of the dataset or the way it was collected and preprocessed/cleaned/labeled that might impact future uses?} \textit{For example, is there anything that a future user might need to know to avoid uses that could result in unfair treatment of individuals or groups (e.g., stereotyping, quality of service issues) or other undesirable harms (e.g., financial harms, legal risks) If so, please provide a description. Is there anything a future user could do to mitigate these undesirable harms?}

\begin{itemize}
\item As this data stems from the greater internet, it mirrors the broader biases of society in the period of its collection. Biases in subpopulation depiction (eg. correlation between gender and jobs), violence, and nudity (for which we provide safety tags) might create harmful outcomes for those a model might be applied to. For this reason this dataset should not be used to make a decision surrounding people.
\end{itemize}

\item \textbf{Are there tasks for which the dataset should not be used?} \textit{If so, please provide a description.}

\begin{itemize}
\item Due to the known biases of the dataset, under no circumstance should any models be put into production using the dataset as is. It is neither safe nor responsible. As it stands, the dataset should be solely used for research purposes in its uncurated state.
\item Likewise, this dataset should not be used to aid in military or surveillance tasks. 
\end{itemize}

\item \textbf{Any other comments?}

\begin{itemize}
\item No.
\end{itemize}

\subsection{Distribution}

\item \textbf{Will the dataset be distributed to third parties outside of the entity (e.g., company, institution, organization) on behalf of which the dataset was created?} \textit{If so, please provide a description.}

\begin{itemize}
\item Yes, the dataset will be open-source.
\end{itemize}

\item \textbf{How will the dataset be distributed (e.g., tarball on website, API, GitHub)?} \textit{Does the dataset have a digital object identifier (DOI)?}

\begin{itemize}
\item The data will be available through Huggingface datasets.
\end{itemize}

\item \textbf{When will the dataset be distributed?}

\begin{itemize}
\item 31/03/2022 and onward.
\end{itemize}

\item \textbf{Will the dataset be distributed under a copyright or other intellectual property (IP) license, and/or under applicable terms of use (ToU)?} \textit{If so, please describe this license and/or ToU, and provide a link or other access point to, or otherwise reproduce, any relevant licensing terms or ToU, as well as any fees associated with these restrictions.}

\begin{itemize}
\item CC-BY-4.0
\end{itemize}

\item \textbf{Have any third parties imposed IP-based or other restrictions on the data associated with the instances?} \textit{If so, please describe these restrictions, and provide a link or other access point to, or otherwise reproduce, any relevant licensing terms, as well as any fees associated with these restrictions.}

\begin{itemize}
\item LAION owns the metadata and release as CC-BY-4.0.
\item We do not own the copyright of the images or text.
\end{itemize}

\item \textbf{Do any export controls or other regulatory restrictions apply to the dataset or to individual instances?} \textit{If so, please describe these restrictions, and provide a link or other access point to, or otherwise reproduce, any supporting documentation.}

\begin{itemize}
\item No.
\end{itemize}

\item \textbf{Any other comments?}

\begin{itemize}
\item No.
\end{itemize}

\subsection{Maintenance}

\item \textbf{Who will be supporting/hosting/maintaining the dataset?}

\begin{itemize}
\item Huggingface will support hosting of the metadata.
\item The Eye supports hosting of the embeddings and backups of the rest.
\item LAION will maintain the samples distributed.
\end{itemize}

\item \textbf{How can the owner/curator/manager of the dataset be contacted (e.g., email address)?}

\begin{itemize}
\item \url{https://laion.ai/dataset-requests/}
\end{itemize}

\item \textbf{Is there an erratum?} \textit{If so, please provide a link or other access point.}

\begin{itemize}
\item There is no erratum for our initial release. Errata will be documented as future releases on the dataset website.
\end{itemize}

\item \textbf{Will the dataset be updated (e.g., to correct labeling errors, add new instances, delete instances)?} \textit{If so, please describe how often, by whom, and how updates will be communicated to users (e.g., mailing list, GitHub)?}

\begin{itemize}
\item LAION-5B will not be updated. However a future LAION-streamed-from-CC may exist for updates. Specific samples can be removed on request.
\end{itemize}

\item \textbf{If the dataset relates to people, are there applicable limits on the retention of the data associated with the instances (e.g., were individuals in question told that their data would be retained for a fixed period of time and then deleted)?} \textit{If so, please describe these limits and explain how they will be enforced.}

\begin{itemize}
\item People may contact us at the \href{https://laion.ai/#contact}{LAION website} to add specific samples to a blacklist.
\end{itemize}

\item \textbf{Will older versions of the dataset continue to be supported/hosted/maintained?} \textit{If so, please describe how. If not, please describe how its obsolescence will be communicated to users.}

\begin{itemize}
\item We will continue to support LAION-400M.
\end{itemize}

\item \textbf{If others want to extend/augment/build on/contribute to the dataset, is there a mechanism for them to do so?} \textit{If so, please provide a description. Will these contributions be validated/verified? If so, please describe how. If not, why not? Is there a process for communicating/distributing these contributions to other users? If so, please provide a description.}

\begin{itemize}
\item Unless there are grounds for significant alteration to certain indexes, extension of the dataset will be carried out on an individual basis.
\end{itemize}

\item \textbf{Any other comments?}

\begin{itemize}
\item No.
\end{itemize}

\end{enumerate}

%% file: sections/dataset_preparation.tex
After processing and filtering common crawl, 5B of image url/text samples are available.
Here we provide an overview of all the steps necessary to combine the full dataset.
\begin{enumerate}
\item Downloading the data as webdataset with distributed img2dataset
\item Computing Vit-L/14 embeddings with distributed clip-inference
\item Computing a KNN index from these embeddings using autofaiss
\item Computing additional tags (NSFW and watermark) using CLIP embeddings
\end{enumerate}

%% file: sections/technical_curation.tex

\subsection{Distributed img2dataset}
We developed img2dataset library to easily download, resize, and store images and captions in the webdataset format.\footnote{\href{https://github.com/rom1504/img2dataset}{https://github.com/rom1504/img2dataset}} This allows to download 100 million images from our list of URLs in 20 hours with a single node (1Gbps connection speed, 32GB of RAM, an i7 CPU with 16 cores), allowing anyone to obtain the whole dataset or a smaller subset.

For LAION-5B we introduced a distributed mode for this tool, allowing to download the 5B samples in a week using 10 nodes.
see \footnote{\href{https://github.com/rom1504/img2dataset/blob/main/dataset_examples/laion5B.md}{https://github.com/rom1504/img2dataset/blob/main/dataset\_examples/laion5B.md}} and
\footnote{\href{https://github.com/rom1504/img2dataset/blob/main/examples/distributed_img2dataset_tutorial.md}{https://github.com/rom1504/img2dataset/blob/main/examples/distributed\_img2dataset\_tutorial.md}}

\subsection{Distributed CLIP inference}
\label{sec:appendix_dist_inference}
From these images, the CLIP retrieval inference tool \footnote{\href{https://github.com/rom1504/clip-retrieval}{https://github.com/rom1504/clip-retrieval}} was used to compute ViT-L/14 embeddings, allowing for a better analysis capacity of the data. In particular a distributed mode \footnote{\href{https://github.com/rom1504/clip-retrieval/blob/main/docs/distributed\_clip\_inference.md}{https://github.com/rom1504/clip-retrieval/blob/main/docs/distributed\_clip\_inference.md}} made it possible to compute these embeddings in a week using 32 NVIDIA A100s: this larger CLIP model can only be computed at a speed of 312 sample/s per gpu, compared to 1800 sample/s for ViT-B/32.

The resulting embeddings are available for everyone to use for clustering, indexing, linear inference.

\subsection{Distributed indexing}
\label{sec:appendix_dist_indexing}
We then used these 9TB of image embeddings to build a large PQ128 knn index using the autofaiss tool \footnote{\href{https://github.com/criteo/autofaiss}{https://github.com/criteo/autofaiss}}. To make this run faster, a distributed mode is available \footnote{\href{https://github.com/criteo/autofaiss/blob/master/docs/distributed/distributed_autofaiss.md}{https://github.com/criteo/autofaiss/blob/master/docs/distributed/distributed\_autofaiss.md}}

\subsection{Integration in the search UI}
\label{sec:appendix_ui_search}
In order to demonstrate the value of this data, we integrated this index into the \footnote{\href{https://knn5.laion.ai}{https://knn5.laion.ai}} UI. It is powered by the code called clip back at \footnote{\href{https://github.com/rom1504/clip-retrieval}{https://github.com/rom1504/clip-retrieval}}
The knn index is 800GB and the metadata (url and captions) as well, so memory mapping is used for both in order to use no RAM, only a SSD drive of that capacity is required.

\subsection{Specialized NSFW image content tagging}
\label{sec:appendix_nsfwtagging}
We applied various tagging to the content of LAION 5B. Among other contents, we tagged images with pornographic or sexualized content (referred to as NSFW). To ensure all implementations related to LAION-5B are open-source, we refrained from using existing commercial solutions.

%
In particular, we first trained an EfficientNetV2-based classifier. However, then moved to a simple MLP based on OpenAI's CLIP/L-14. 
To this end, we created a training dataset by retrieving images from the previous LAION-400M dataset which are close in the CLIP embedding space to various keywords related to the five categories: ``neutral", ``drawing'', ``porn'', ``hentai'' or ``sexy''. 
Additionally, we added SFW images from the Wikiart \footnote{\href{https://www.wikiart.org}{https://www.wikiart.org}} and Danbooru datasets \footnote{\href{https://www.gwern.net/Danbooru2021}{https://www.gwern.net/Danbooru2021}} to the ``drawing'' category and NSFW images from Danbooru to the ``hentai'' category. 

Following this procedure, we obtained over 682K images from the five classes ``drawing'' (39026), ``hentai'' (28134), ``neutral'' (369507), ``porn'' (207969) and ``sexy'' (37914).
Using this data we trained a detector for these five categories by finetuning an ImageNet-1k pretrained EfficientNet-V2-B02 model. \footnote{Code may be found at: \href{https://github.com/LAION-AI/LAION-SAFETY}{https://github.com/LAION-AI/LAION-SAFETY}}
To use this image classifier as a binary SFW - NSFW classifier, we consider images from the classes ``drawing'' and ``neutral'' as SFW and ``hentai'', ``porn'' and ``sexy'' as NSFW.
To measure the performance of this model, we created a test dataset with 1000 images from each category and manually inspected it, to make sure all test images where correctly annotated. 
Our EfficientNet-V2-B02 image classifier predicted 96,45\% of the true NSFW correctly as NSFW and discards 7,96\% of the SFW images incorrectly as NSFW.

\subsection{Further inappropriate content tagging}
\label{sec:appendix_q16tagging}
\begin{figure}[t]
    \centering
    \includegraphics[width=0.9\linewidth]{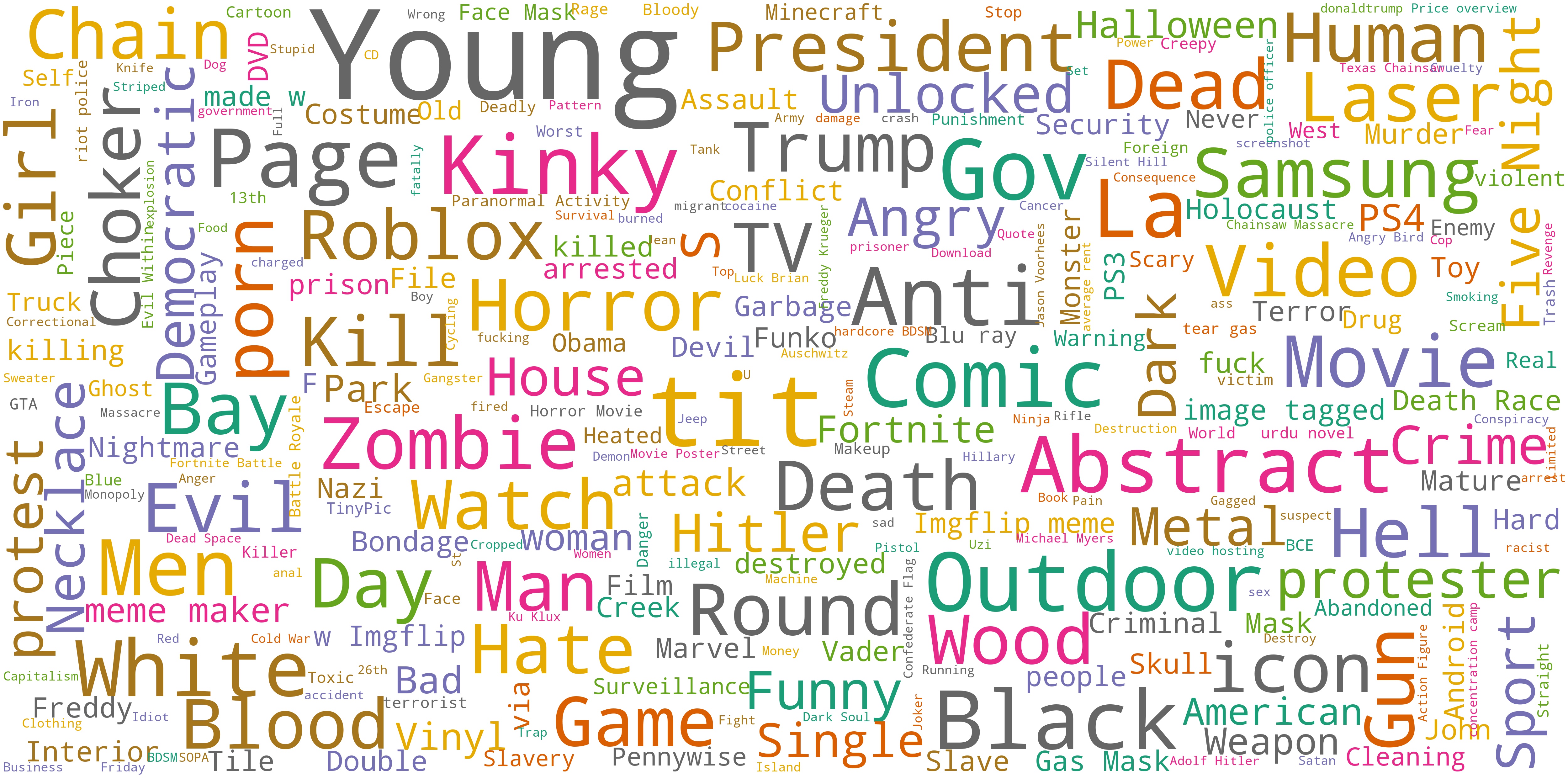}
    \caption{Word cloud based on \cite{schramowski2022can} documenting the potentially inappropriate image content of the LAION-5B subset which contains text in English language. Provided alternative text is used as text description of the images.
    Word size is proportional to the word counts and rank in descriptions corresponding to the inappropriate image set.}
    \label{fig:q16_inappropriate_content_doc}
\end{figure}

Further, we used the Q16 documentation pipeline \cite{schramowski2022can} to document the broad range of identified potentially inappropriate concepts contained, cf.~Sec.~\ref{subsec:safety_tagging} for details. Fig.~\ref{fig:q16_inappropriate_content_doc} shows the most frequent identified concepts following this procedure. One can see that in a lot of cases these images show humans (cf. concepts \textit{human, people, man, woman}). Further, one main concept is pornographic content (e.g. \textit{porn, bondage, kinky, bdsm}). Additionally, most frequent present concepts are, among other concepts, \textit{weapons, violence, terror, murder, slavery, racism} and \textit{hate}. Note that also content surrounding \textit{halloween} (\textit{costume, halloween, zombie}) and art or media such as \textit{movie}s, \textit{game}s and \textit{comic}s are potentially tagged, depending on the displayed content. Further filtering depends highly on the use-case and users' opinions.

\subsection{Watermark and safety inference}

\begin{figure}[!tb]
    \centering
\subfloat{{\includegraphics[width=\linewidth]{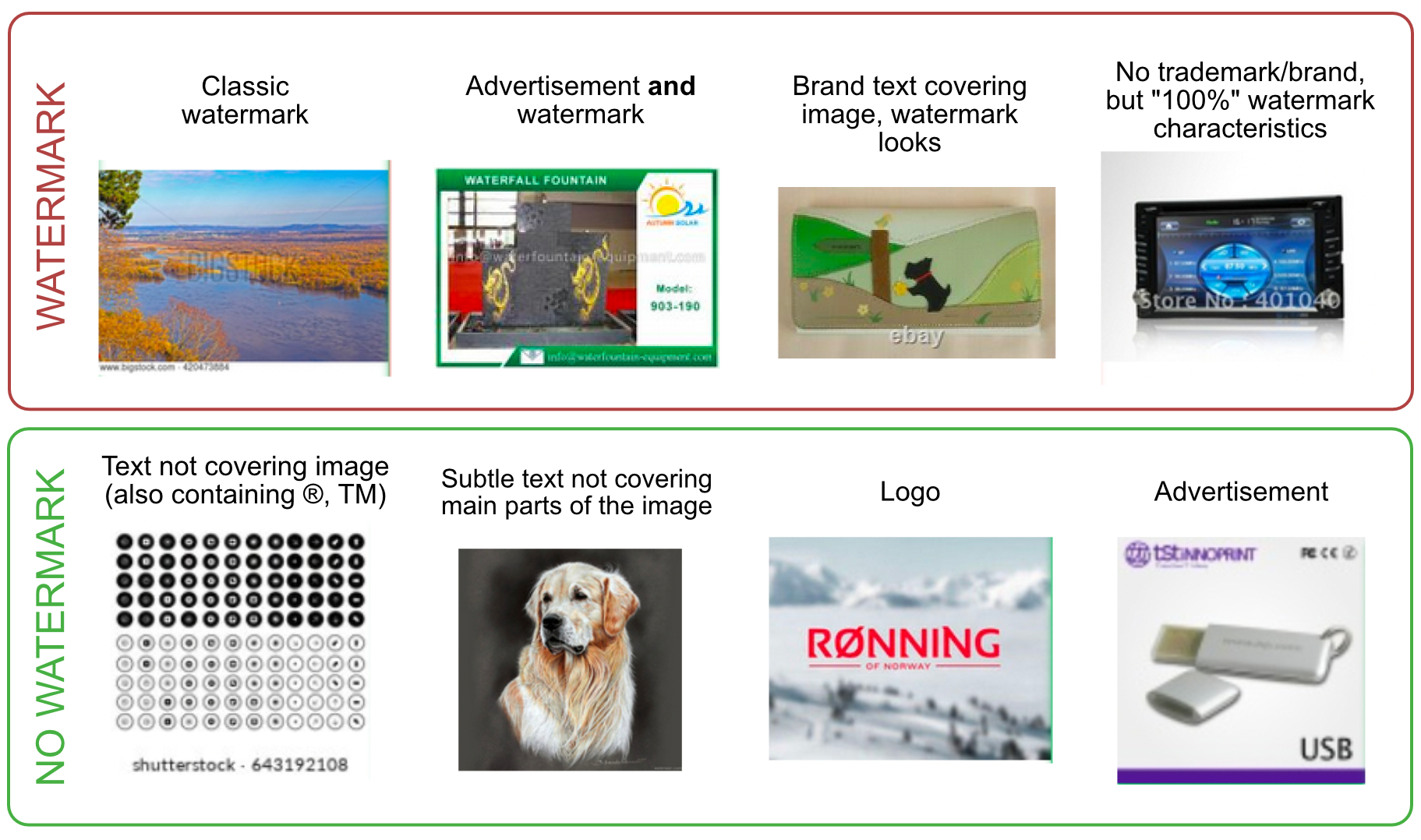}}}\\
    \caption{\textbf{Watermark test set annotation examples.} Criteria for LAION-5B sample annotation for watermark (top row) and non-watermark (bottom row) images.} 
    \label{watermark_annotation}
\end{figure}

Finally, we wanted to let user the ability to remove unsafe examples, and watermarked examples. To do that we collected training and test sets. The training set was augmented with examples retrieved from the KNN index, while the test set samples were selected to represent well the dataset distribution but were all manually annotated. \ref{watermark_annotation}

The inference is done using the embedding-reader\footnote{\href{https://github.com/rom1504/embedding-reader}{https://github.com/rom1504/embedding-reader}} module.

These tags were then integrated in the UI, allowing everyone to observe that the safety tags indeed filter out almost all the unsafe results, and giving confidence that training a generative model on this data will not result in unexpectedly unsafe images.

%% file: sections/experiment_samples.tex
\begin{figure}[tb!]
    \centering
    \includegraphics[width=\linewidth]{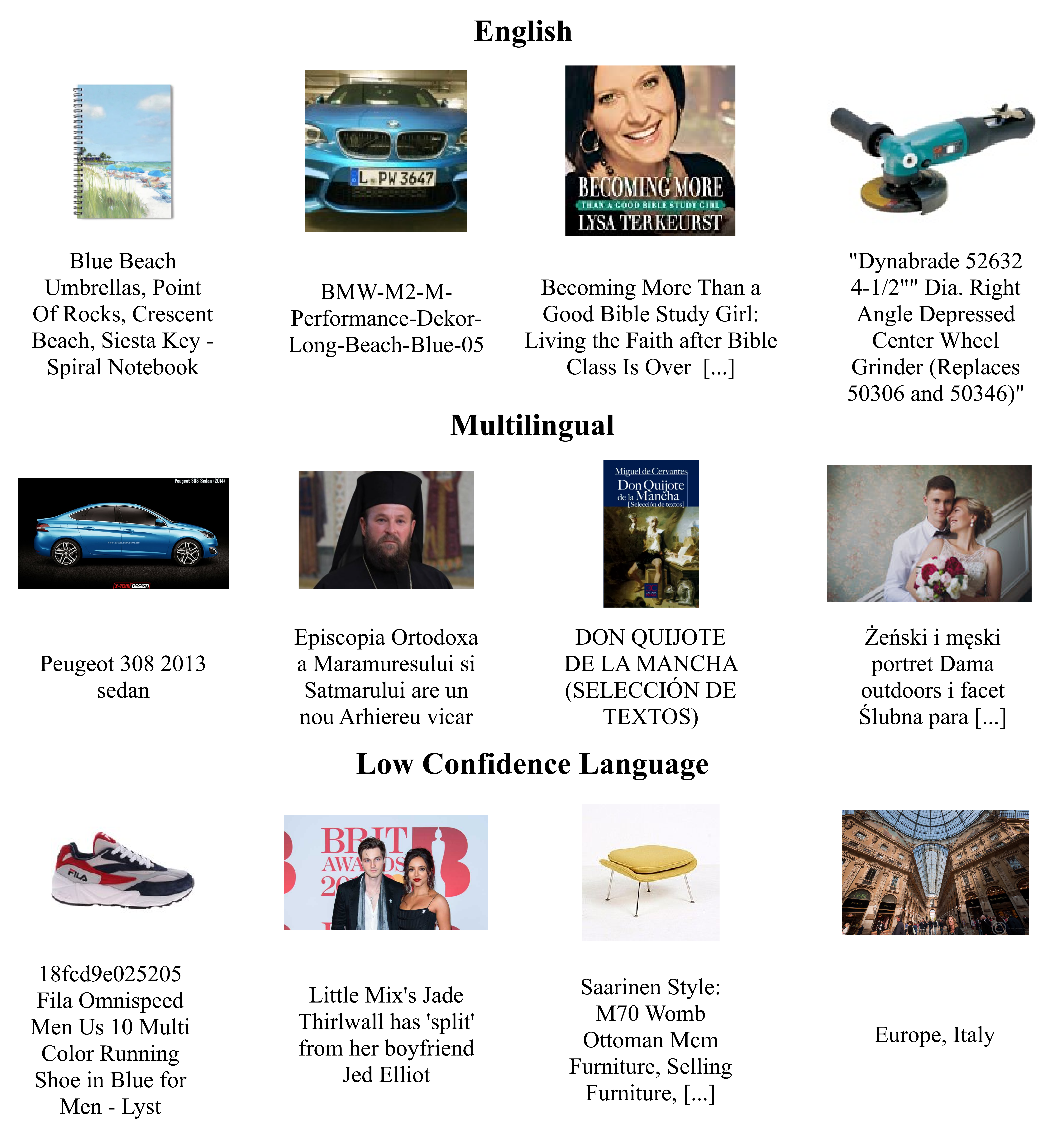}
    \caption{\textbf{LAION-5B random examples from all subsets.} We take the first 4 SFW samples from each of the 3 randomly shuffled LAION-5B subsets. We present the image and its associated caption.}
    \label{fig:random_examples}
\end{figure}

Here, we present samples from the dataset and some distribution statistics to aid in understanding the dataset. 
In Figure \ref{fig:random_examples}, we randomly select 4 samples from each of the 3 LAION-5B subsets. As can be seen, the language classifier seems to have low confidence with names, identifying numbers, and short form text. An important future line of work will be to improve the language classifier.

\begin{figure}[tb!]
\begin{floatrow}
\ffigbox{%
  \centering
  \includegraphics[width=0.48\textwidth]{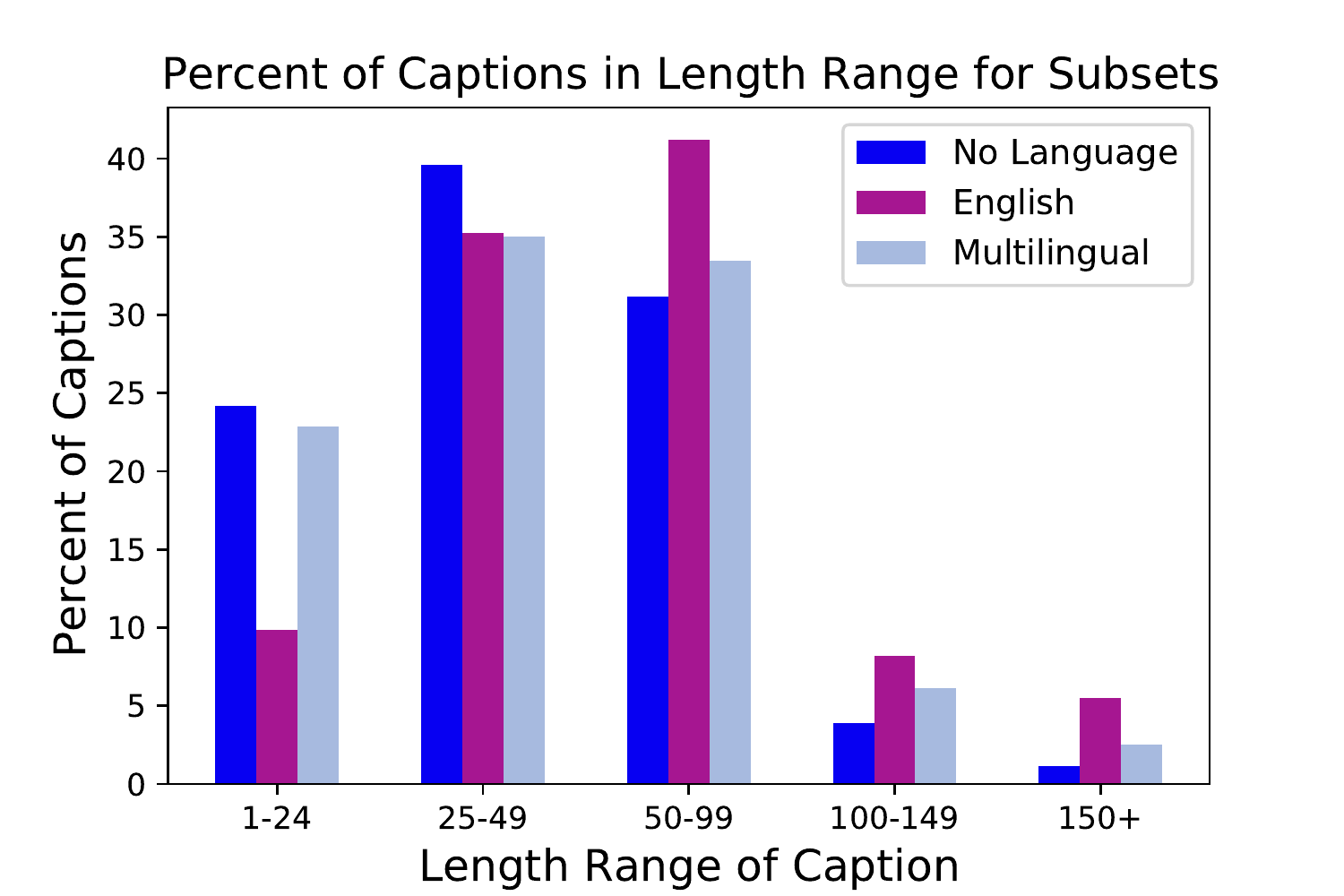}%
}{%
\newline
\newline
  \caption{\textbf{Caption Character Length.} Each of the LAION-5B subsets contains similar frequencies and exhibit a right skew.}%
  \label{fig:length}
}

\ffigbox{%
\centering
\includegraphics[width=0.48\textwidth]{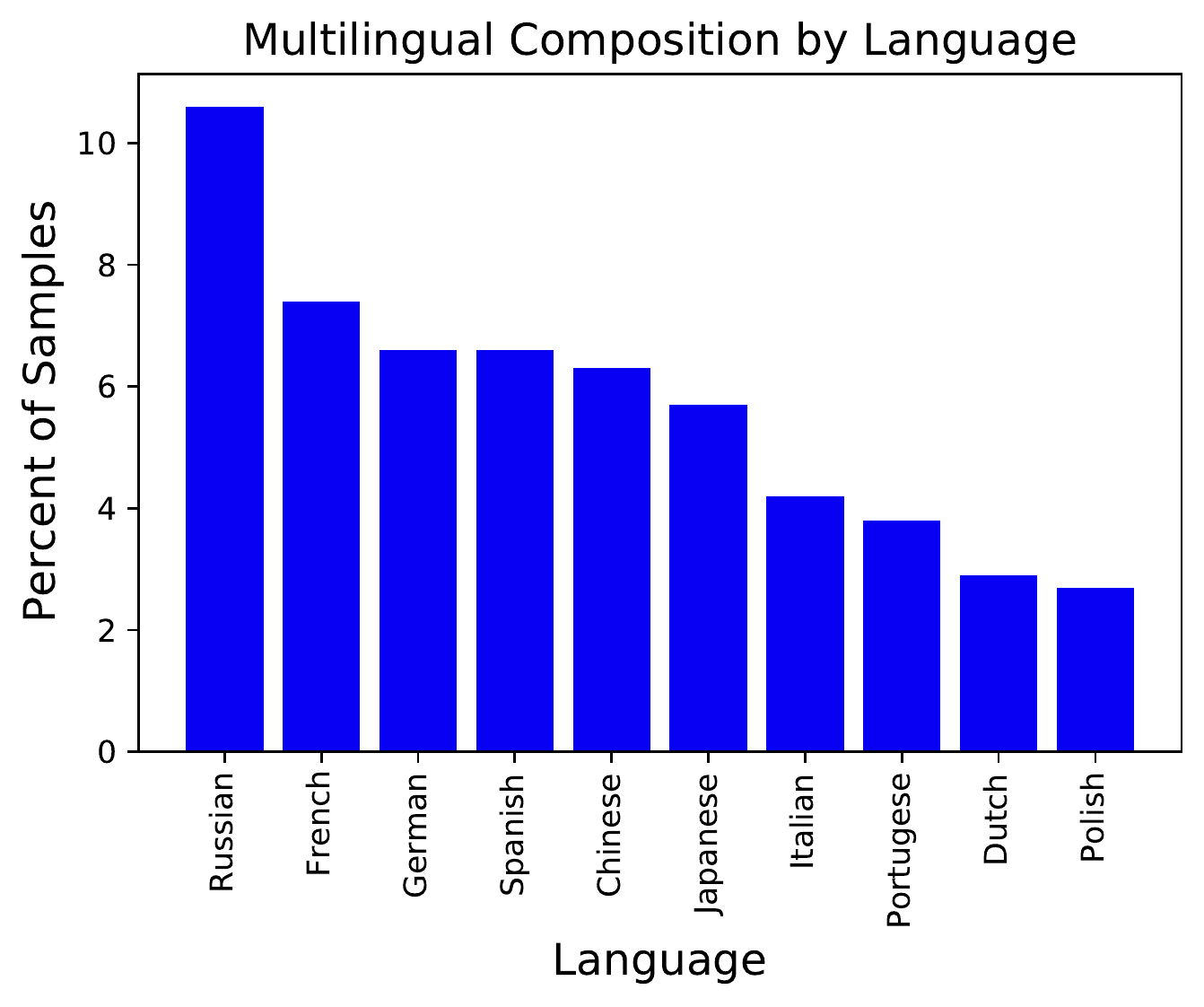}%
}{%
  \caption{\textbf{Multilingual Language Frequency.} The 10 most frequent languages seem to be largely of European and East Asian origin.}%
  \label{fig:language_frequency}
}
\end{floatrow}
\end{figure}

To comprehend the dataset beyond visual examples, we may look at statistics collected about the distribution. Figure \ref{fig:length} gives an overview of the caption length amongst all subsets. Additionally, Figure \ref{fig:language_frequency} describes the frequency of languages within the multilingual subset. The 10 most frequent languages compose 56\% of the multilingual dataset.


%% file: sections/experiments_appendix.tex
We provide details about experiments that were
done to reproduce CLIP~\cite{radford2021learning} using LAION (400M, 2B-en) subsets. In addition, we document all experimental results on both zero-shot classification using the VTAB+ suite and retrieval.

\subsection{Training Details}
\label{sec:appendix_experiments_training_details}


We used distributed data parallel training (using PyTorch DDP) to train models on multiple NVIDIA A100 GPUs. Training was done using the InfoNCE loss like in \cite{radford2021learning}. We used Adam with decoupled weight regularization (i.e., AdamW) as an optimizer, with $\beta_1=0.9$ and $\beta_2=0.98$ for all models. We used a linear warmup followed by a cosine decay schedule. For regularization we used the same weight decay of $0.2$ for all the models. 
Details about different architectures that were used are provided in Tab.~\ref{table:openclip_architectures}.
Training hyper-parameters and resources used are provided in Tab.~\ref{table:openclip_hyper_parameters}.

\subsection{Distributed Training and InfoNCE Loss}
\label{sec:appendix_distributed_infonce}

To properly deal with global batch for contrastive InfoNCE loss in distributed setting, we need additional communication between GPU workers to compute the loss and the gradients for all positive and negative sample pairs correctly. In each worker, we gather all image and text embeddings from the other workers, and use them as negative examples for each image-text pair in the mini-batch.

A naive implementation of InfoNCE involves materializing a very large $N\times N$ matrix, $N$ being the global batch size. For $N=32768$, the matrix occupies a hefty 8 GB in float32. To remedy this, we use a formulation of the loss like OpenAI ~\cite{radford2021learning} where redundant operations are sharded to local devices while maintaining correct global gradients. This successfully overcomes a significant scaling issue and achieves a memory complexity that scales linearly with global batch size by only materializing 2 matrices of size $n\times N$, $n$ being local batch size per GPU. By turning memory complexity from $\mathcal{O}(N^2)$ into $\mathcal{O}(nN)$, we slash memory overhead due to scaling from GBs down to MBs.

\begin{table}[ht]
\rowcolors{2}{light-light-gray}{white}
\begin{tabular}{@{}lllllll@{}}
\toprule
\textbf{Name} & \textbf{Width}  & \textbf{Embed Dim} & \textbf{Depth} & \textbf{Res.} & \textbf{Acts.} & \textbf{Params} \\ \midrule
ViT-B/32      & 768 / 512                 & 512                & 12 / 12        & 224x224             & 10 M                           & 151 M                     \\
ViT-B/16      & 768 / 512                 & 512                & 12 / 12        & 224x224             & 29 M                           & 150 M                     \\
ViT-B/16+     & 896 / 640                 & 640                & 12 / 12        & 240x240             & 40 M                           & 208 M                     \\ 
ViT-L/14     & 1024 / 768                 & 768                & 24 / 12        & 224x224             & 97 M                           & 428 M                     \\ 
\bottomrule
\end{tabular}
\caption{Hyper-parameters of different architectures we used for reproducing CLIP models. \textbf{Acts} refers to the number of activations in millions, while \textbf{Params} refers to the number of parameters in millions. All entries in the form of A / B denote image and text parameters respectively.} 
\label{table:openclip_architectures}
\end{table}

\begin{table}[ht]
\centering
\rowcolors{2}{light-light-gray}{white}
\begin{tabular}{@{}lllclll@{}}
\toprule
\textbf{Model (data size)} & \textbf{BS. (global)} & \textbf{\#GPUs} & \textbf{LR.} & \textbf{Warm.} & \textbf{Ep}. & \textbf{Time (hrs.)} \\ \midrule
B/32 (400M)                & 256 (32768)        & 128    & $5\text{e-}4$          & 2K         & 32     & 36                    \\
B/32 (2B)                  & 416 (46592)        & 112    & $5.5\text{e-}4$        & 10K        & 16     & 210                  \\
B/16 (400M)                & 192 (33792)        & 176    & $5\text{e-}4$          & 2K         & 32     & 61                    \\
B/16+(400M)                & 160 (35840)        & 224    & $7\text{e-}4$          & 5K         & 32     & 61                    \\
L/14 (400M)                & 96 (38400)         & 400    & $6\text{e-}4$          & 5K         & 32     & 88                     \\ \bottomrule
\end{tabular}
\caption{Training hyper-parameters and resources used to reproduce CLIP~\cite{radford2021learning} models on LAION 400M and 2B subsets. Note that \textbf{BS} refer to batch size per GPU worker (with \textbf{global} the corresponding global batch size), \textbf{LR}  to base learning rate, \textbf{Warm} to the total number of warmup steps, \textbf{Ep} to the total number of training epochs, and \textbf{Time} to total training time in hours.}
\label{table:openclip_hyper_parameters}
\end{table}




\subsection{Detailed Results \& Further Analysis}
\label{sec:appendix_experiments_evaluation_details}

In this section we present all zero-shot classification results on VTAB+ as well as retrieval results.
In Tab.~\ref{table:vtab_plus_dataset_info}, we describe the datasets that are used in VTAB+.
For zero-shot classification, we collected prompts and class names from prior works~\cite{radford2021learning,zhai2021lit}
and made them available in our benchmark repository\footnote{\href{https://github.com/LAION-AI/CLIP_benchmark/blob/d4e52e8655c0b943c282552a3375f59d2d40c2bd/clip_benchmark/datasets/builder.py\#L474}{https://github.com/LAION-AI/CLIP\_benchmark}}. In Tab.~\ref{table:zeroshot_detailed},  we show zero-shot top-1 classification accuracy (\%) on VTAB+ datasets.
Tables \ref{table:flickr} and \ref{table:mscoco} depict retrieval results on Flickr30K\cite{young-etal-2014-image} and MSCOCO \cite{lin2014microsoft}.

\begin{table}[!h]
\centering
\rowcolors{2}{light-light-gray}{white}
\begin{tabular}{@{}llll@{}}
\toprule
\bf{Dataset} & \bf{Abbr.(Tab.~\ref{table:zeroshot},~\ref{table:zeroshot_detailed})}  & \bf{Test size} & \bf{\#Classes}\\\midrule
ImageNet-1k & INet & 50,000 &1,000\\
ImageNet-v2 & INet-v2 & 10,000 &1,000\\
ImageNet-R & INet-R & 30,000 &200\\
ImageNet Sketch & INet-S & 50,889 &1,000\\
ObjectNet & ObjNet & 18,574 &113\\
ImageNet-A & INet-A & 7,500 &200\\
CIFAR-10 & - & 10,000 &10\\
CIFAR-100 & - & 10,000 &100\\
MNIST & - & 10,000 &10\\
Oxford Flowers 102 & Flowers102 & 6,149 &102\\
Stanford Cars & Cars & 8,041 &196\\
SVHN & - & 26,032 &10\\
Facial Emotion Recognition 2013 & FER2013 & 7,178 &7\\
RenderedSST2 & - & 1,821 &2\\
Oxford-IIIT Pets & Pets & 3,669 &37\\
Caltech-101 & - & 6,085 &102\\
Pascal VOC 2007 Classification & VOC2007-Cl & 14,976 &20\\
SUN397 & - & 108,754 &397\\
FGVC Aircraft & - & 3,333 &100\\
Country211 & - & 21,100 &211\\
Describable Textures & DTD & 1,880 &47\\
GTSRB & - & 12,630 &43\\
STL10 & - & 8,000 &10\\
Diabetic Retinopathy & Retino & 42,670 &5\\
EuroSAT & - & 5,400 &10\\
RESISC45 & - & 6,300 &45\\
PatchCamelyon & PCAM & 32,768 &2\\
CLEVR Counts & - & 15,000 &8\\
CLEVR Object Distance & CLEVR Dist & 15,000 &6\\
DSPRITES Orientation & DSPRITES Orient & 73,728 &40\\
DSPRITES Position & DSPRITES pos & 73,728 &32\\
SmallNORB Elevation & SmallNORB Elv & 12,150 &9\\
SmallNORB Azimuth & SmallNORB Azim & 12,150 &18\\
DMLAB & - & 22,735 &6\\
KITTI closest vehicle distance & KITTI Dist & 711 &4\\
\bottomrule\end{tabular}
\caption{Datasets used for zero-shot classification evaluation (VTAB+).}
\label{table:vtab_plus_dataset_info}
\end{table}

\begin{figure}[tb]
    \centering
    
    \includegraphics[width=0.7\linewidth]{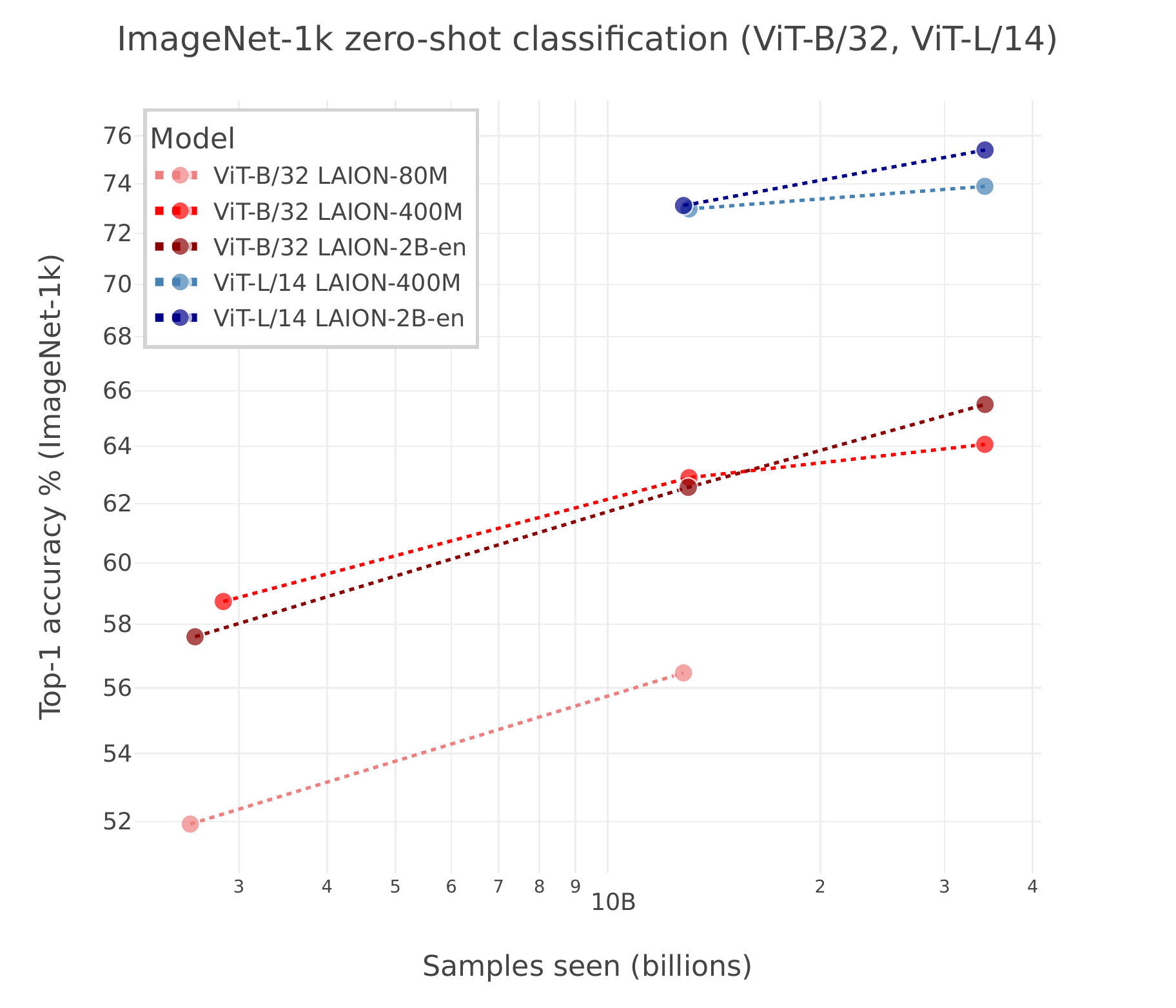}
    
    \caption{
        ViT-B/32 and ViT-L/14 additional experiments where we vary the amount compute (3B, 13B, and 34B images seen) and LAION subset size (80M, 400M, 2B). We evaluate the models on zero-shot Imagenet-1k classification. Seeing same number of samples on larger data scale leads consistently to better zero-shot transfer performance, when investing enough into training compute. 
    }
    \label{fig:data_scale}
\end{figure}

\paragraph{Effect of data scale.} We observe similar or better results on most datasets when using the larger LAION-2B-en instead of LAION-400M. Exceptions are on some datasets with specialized domains (e.g., Diabetic Retinopathy, PatchCamelyon) or in structured tasks (see corresponding paragraph below). To demonstrate the importance of the data scale for the quality of the pre-trained models, we conduct a series of experiments where we vary both data scale (LAION-80M, LAION-400M and LAION-2B) and amount of training compute measured in samples seen (3B, 13B and 34B). We observe that when investing enough into training compute, seeing same number of samples on larger data scale leads consistently to better zero-shot transfer performance measured on ImageNet-1k. This is valid for both smaller B/32 and larger L/14 model scales. For instance, models pre-trained on LAION-2B outperform there significantly models pre-trained on LAION-400M, when using same large compute training budget of 34B samples seen (see Fig.~\ref{fig:data_scale} and Tab.~\ref{table:data_scale}). We conclude from these findings that extending dataset scale all the way up towards LAION-2B is indeed important for obtaining stronger zero-shot transfer performance, given sufficiently large compute for training.

\begin{figure}[tb]
    \centering
    \includegraphics[width=0.5\textwidth]{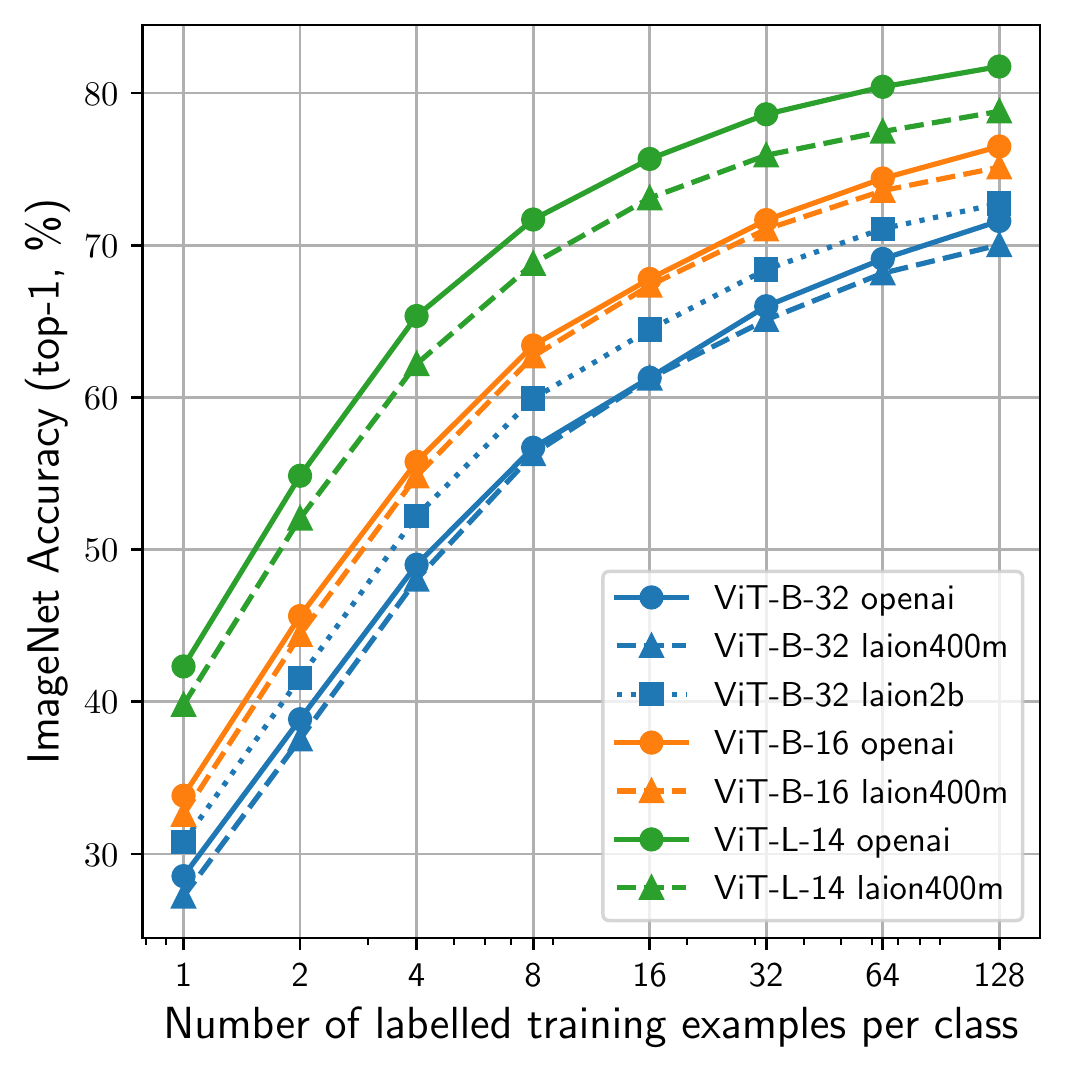}
    \caption{Evaluating few-shot linear probe performance on ImageNet. We evaluate i) models trained on various LAION subsets and ii) the original CLIP models. Models trained on LAION show similar transfer performance to those trained by OpenAI. Also evident is clear effect of model or data scale on transfer across few-shot conditions.}
    \label{fig:probes-in}
\end{figure}
\begin{figure}[tb]
    \centering
    \includegraphics[width=\textwidth]{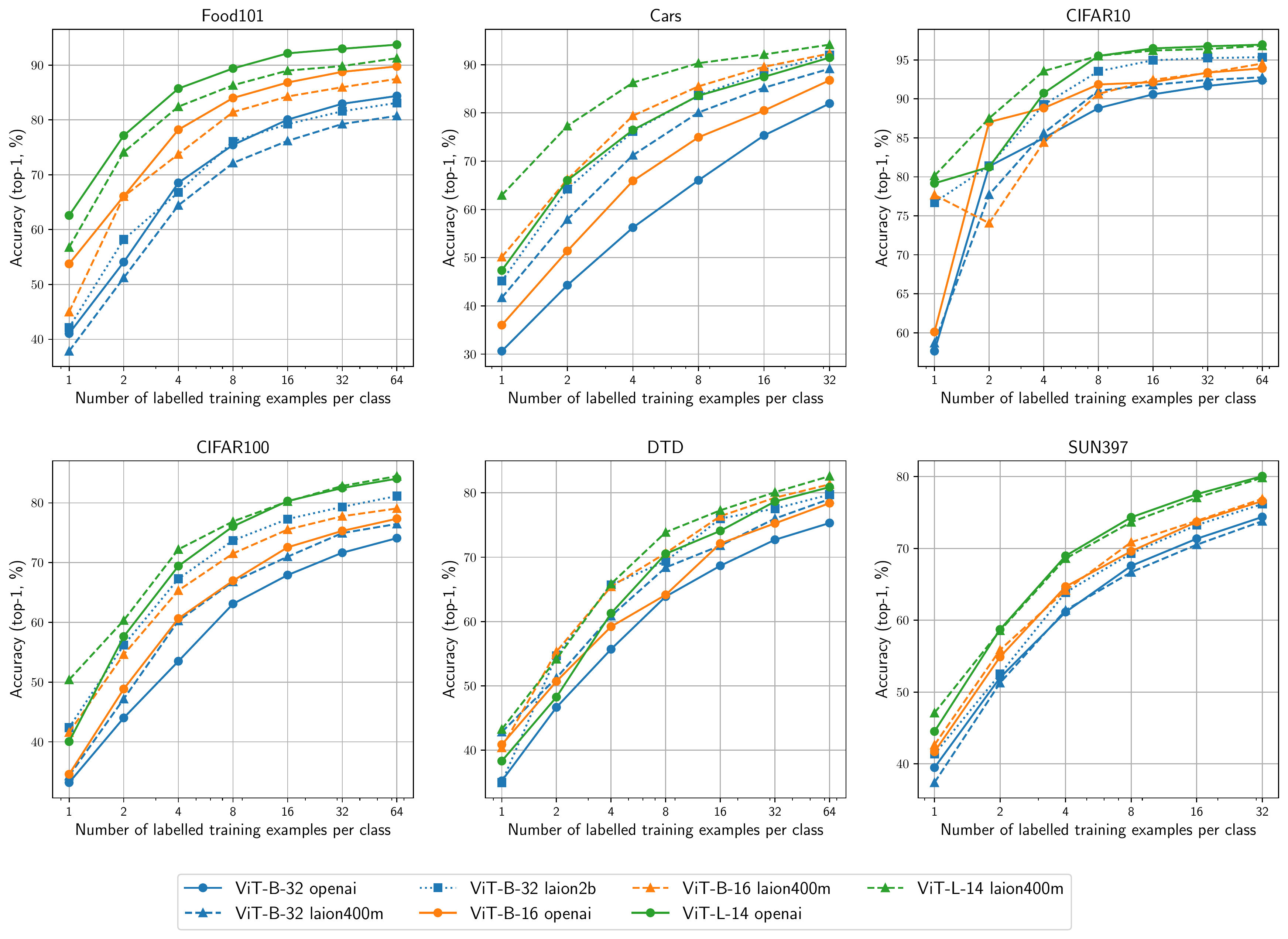}
    \caption{Evaluating few-shot linear probe performance on 6 datasets commonly used to benchmark transfer~\cite{kornblith2019better}. We evaluate i) models trained on various LAION subsets and ii) the original CLIP models. We evaluate performance on Food101~\cite{food101}, Cars~\cite{cars}, CIFAR-10 \& 100~\cite{krizhevsky2009learning}, DTD~\cite{dtd} and SUN397~\cite{sun397}.}
    \label{fig:probes-transfer}
\end{figure}

\paragraph{Few-shot transfer: comparison to CLIP and effect of scale.}
To examine the quality of the learned representations, we evaluate few-shot linear probe performance on seven datasets commonly used to benchmark transfer performance.
The results are presented in Figures~\ref{fig:probes-in} and~\ref{fig:probes-transfer}.
Figure~\ref{fig:probes-in} displays few-shot performance on ImageNet~\cite{deng2009imagenet} while Figure~\ref{fig:probes-transfer} displays few-shot performance on Food101~\cite{food101}, Cars~\cite{cars}, CIFAR-10 \& 100~\cite{krizhevsky2009learning}, DTD~\cite{dtd} and SUN397~\cite{sun397}.
In addition to evaluating models trained on subsets of LAION, we also compare with the CLIP models of Radford \textit{et al.}~\cite{radford2021learning}.
Overall we observe that the models trained on LAION achieve similar transfer performance to those trained by OpenAI.
Moreover, we observe that performance increases with more data (i.e., B/32 2B outperforms B/32 400M) and larger models.


\paragraph{ImageNet-A} In ImageNet-A~\cite{hendrycks2021natural} (noted INet-A), we observe large differences between CLIP WIT and LAION models, e.g.\ a difference of 24.3\% on ViT-L/14.
We note that INet-A design and data collection is quite different from other ImageNet distribution shifts datasets, as the images were specifically selected to be adversarial for a ResNet-50 pre-trained on ImageNet-1k. Although we do not have yet an explanation for the observed discrepancies and it would be interesting to understand why LAION models are worse than CLIP WIT, it is not clear whether improvements in INet-A are generalizable, as the dataset is based on adversarial images specific to a pre-trained model (ResNet-50).

\begin{table}[tb]
\rowcolors{2}{light-light-gray}{white}
\small

\begin{tabular}{@{}lclll@{}}
\toprule
Model                     & Samples seen & LAION-80M & LAION-400M & LAION-2B-en \\ \midrule
\textbf{ViT-B/32} & 3B           & 51.93     & 58.73      & 57.60       \\
                          & 13B          & 56.46     & 62.90       & 62.56       \\
                          & 34B          & -         & 64.07      & 65.50        \\ \midrule
\textbf{ViT-L/14}         & 13B          & -         & 72.98      & 73.12           \\
                          & 34B          & -         & 73.90       & 75.40        \\ \bottomrule 
\end{tabular}
\caption{
    ViT-B/32 and ViT-L/14 additional experiments where we vary the amount compute (3B, 13B, and 34B images seen) and LAION subset size (80M, 400M, 2B). We evaluate the models on zero-shot Imagenet-1k classification. When investing enough into training compute, seeing same number of samples on larger data scale leads consistently to better zero-shot transfer performance measured on ImageNet-1k.
}
\label{table:data_scale}
\end{table}

\begin{table}[p]
\setlength{\tabcolsep}{3pt}
\footnotesize
\rowcolors{2}{light-light-gray}{white}
\begin{tabular}{@{}lcll|cl|c|cl@{}}
\toprule
&\multicolumn{3}{c|}{B/32}&\multicolumn{2}{c|}{B/16}&\multicolumn{1}{c|}{B/16+}&\multicolumn{2}{c}{L/14}\\
Dataset & \tiny{CLIP WIT}&\tiny{LAION-400M}&\tiny{LAION-2B}&\tiny{CLIP WIT}&\tiny{LAION-400M}&\tiny{LAION-400M}&\tiny{CLIP WIT}&\tiny{LAION-400M}\\\midrule
INet & \color{black}{63.3}&${\color{black}{62.9}}^{\tiny\color{red}\textbf{-0.4}}$&${\color{black}{65.7}}^{\tiny\color{ForestGreen}\textbf{+2.4}}$&\color{black}{68.3}&${\color{black}{67.0}}^{\tiny\color{red}\textbf{-1.3}}$&\color{black}{69.2}&\color{black}{75.6}&${\color{black}{72.8}}^{\tiny\color{red}\textbf{-2.8}}$\\
INet-v2 & \color{black}{56.0}&${\color{black}{55.1}}^{\tiny\color{red}\textbf{-0.9}}$&${\color{black}{57.4}}^{\tiny\color{ForestGreen}\textbf{+1.4}}$&\color{black}{61.9}&${\color{black}{59.6}}^{\tiny\color{red}\textbf{-2.3}}$&\color{black}{61.5}&\color{black}{69.8}&${\color{black}{65.4}}^{\tiny\color{red}\textbf{-4.4}}$\\
INet-R & \color{black}{69.4}&${\color{black}{73.4}}^{\tiny\color{ForestGreen}\textbf{+4.0}}$&${\color{black}{75.9}}^{\tiny\color{ForestGreen}\textbf{+6.5}}$&\color{black}{77.7}&${\color{black}{77.9}}^{\tiny\color{ForestGreen}\textbf{+0.2}}$&\color{black}{80.5}&\color{black}{87.9}&${\color{black}{84.7}}^{\tiny\color{red}\textbf{-3.2}}$\\
INet-S & \color{black}{42.3}&${\color{black}{49.4}}^{\tiny\color{ForestGreen}\textbf{+7.1}}$&${\color{black}{52.9}}^{\tiny\color{ForestGreen}\textbf{+10.6}}$&\color{black}{48.2}&${\color{black}{52.4}}^{\tiny\color{ForestGreen}\textbf{+4.2}}$&\color{black}{54.4}&\color{black}{59.6}&\color{black}{59.6}\\
ObjNet & \color{black}{44.2}&${\color{black}{43.9}}^{\tiny\color{red}\textbf{-0.3}}$&${\color{black}{48.7}}^{\tiny\color{ForestGreen}\textbf{+4.5}}$&\color{black}{55.3}&${\color{black}{51.5}}^{\tiny\color{red}\textbf{-3.8}}$&\color{black}{53.9}&\color{black}{69.0}&${\color{black}{59.9}}^{\tiny\color{red}\textbf{-9.1}}$\\
INet-A & \color{black}{31.6}&${\color{black}{21.7}}^{\tiny\color{red}\textbf{-9.9}}$&${\color{black}{26.1}}^{\tiny\color{red}\textbf{-5.5}}$&\color{black}{49.9}&${\color{black}{33.2}}^{\tiny\color{red}\textbf{-16.7}}$&\color{black}{36.9}&\color{black}{70.8}&${\color{black}{46.5}}^{\tiny\color{red}\textbf{-24.3}}$\\
CIFAR-10 & \color{black}{89.8}&${\color{black}{90.7}}^{\tiny\color{ForestGreen}\textbf{+0.9}}$&${\color{black}{94.0}}^{\tiny\color{ForestGreen}\textbf{+4.2}}$&\color{black}{90.8}&${\color{black}{91.7}}^{\tiny\color{ForestGreen}\textbf{+0.9}}$&\color{black}{92.7}&\color{black}{95.6}&${\color{black}{94.6}}^{\tiny\color{red}\textbf{-1.0}}$\\
CIFAR-100 & \color{black}{64.2}&${\color{black}{70.3}}^{\tiny\color{ForestGreen}\textbf{+6.1}}$&${\color{black}{75.4}}^{\tiny\color{ForestGreen}\textbf{+11.2}}$&\color{black}{66.9}&${\color{black}{71.2}}^{\tiny\color{ForestGreen}\textbf{+4.3}}$&\color{black}{73.8}&\color{black}{75.9}&${\color{black}{77.4}}^{\tiny\color{ForestGreen}\textbf{+1.5}}$\\
MNIST & \color{black}{48.2}&${\color{black}{37.4}}^{\tiny\color{red}\textbf{-10.8}}$&${\color{black}{63.4}}^{\tiny\color{ForestGreen}\textbf{+15.2}}$&\color{black}{51.8}&${\color{black}{66.3}}^{\tiny\color{ForestGreen}\textbf{+14.5}}$&\color{black}{57.0}&\color{black}{76.4}&${\color{black}{76.0}}^{\tiny\color{red}\textbf{-0.4}}$\\
Flowers102 & \color{black}{66.5}&${\color{black}{68.1}}^{\tiny\color{ForestGreen}\textbf{+1.6}}$&${\color{black}{69.0}}^{\tiny\color{ForestGreen}\textbf{+2.5}}$&\color{black}{71.2}&${\color{black}{69.3}}^{\tiny\color{red}\textbf{-1.9}}$&\color{black}{71.1}&\color{black}{79.2}&${\color{black}{75.6}}^{\tiny\color{red}\textbf{-3.6}}$\\
Cars & \color{black}{59.6}&${\color{black}{79.3}}^{\tiny\color{ForestGreen}\textbf{+19.7}}$&${\color{black}{84.4}}^{\tiny\color{ForestGreen}\textbf{+24.8}}$&\color{black}{64.7}&${\color{black}{83.7}}^{\tiny\color{ForestGreen}\textbf{+19.0}}$&\color{black}{84.5}&\color{black}{77.9}&${\color{black}{89.6}}^{\tiny\color{ForestGreen}\textbf{+11.7}}$\\
SVHN & \color{black}{13.4}&${\color{black}{27.7}}^{\tiny\color{ForestGreen}\textbf{+14.3}}$&${\color{black}{38.8}}^{\tiny\color{ForestGreen}\textbf{+25.4}}$&\color{black}{31.3}&${\color{black}{38.5}}^{\tiny\color{ForestGreen}\textbf{+7.2}}$&\color{black}{36.2}&\color{black}{57.0}&${\color{black}{38.0}}^{\tiny\color{red}\textbf{-19.0}}$\\
FER2013 & \color{black}{41.4}&${\color{black}{43.0}}^{\tiny\color{ForestGreen}\textbf{+1.6}}$&${\color{black}{48.1}}^{\tiny\color{ForestGreen}\textbf{+6.7}}$&\color{black}{46.3}&${\color{black}{43.2}}^{\tiny\color{red}\textbf{-3.1}}$&\color{black}{44.5}&\color{black}{50.1}&${\color{black}{50.3}}^{\tiny\color{ForestGreen}\textbf{+0.2}}$\\
RenderedSST2 & \color{black}{58.6}&${\color{black}{52.3}}^{\tiny\color{red}\textbf{-6.3}}$&${\color{black}{54.3}}^{\tiny\color{red}\textbf{-4.3}}$&\color{black}{60.5}&${\color{black}{54.4}}^{\tiny\color{red}\textbf{-6.1}}$&\color{black}{57.9}&\color{black}{68.9}&${\color{black}{56.0}}^{\tiny\color{red}\textbf{-12.9}}$\\
Pets & \color{black}{87.3}&${\color{black}{86.9}}^{\tiny\color{red}\textbf{-0.4}}$&${\color{black}{89.2}}^{\tiny\color{ForestGreen}\textbf{+1.9}}$&\color{black}{89.0}&${\color{black}{89.2}}^{\tiny\color{ForestGreen}\textbf{+0.2}}$&\color{black}{90.3}&\color{black}{93.3}&${\color{black}{91.9}}^{\tiny\color{red}\textbf{-1.4}}$\\
Caltech-101 & \color{black}{81.6}&${\color{black}{83.2}}^{\tiny\color{ForestGreen}\textbf{+1.6}}$&${\color{black}{83.1}}^{\tiny\color{ForestGreen}\textbf{+1.5}}$&\color{black}{82.2}&${\color{black}{83.6}}^{\tiny\color{ForestGreen}\textbf{+1.4}}$&\color{black}{83.2}&\color{black}{83.3}&${\color{black}{84.0}}^{\tiny\color{ForestGreen}\textbf{+0.7}}$\\
VOC2007-Cl & \color{black}{76.4}&${\color{black}{75.8}}^{\tiny\color{red}\textbf{-0.6}}$&${\color{black}{78.8}}^{\tiny\color{ForestGreen}\textbf{+2.4}}$&\color{black}{78.3}&${\color{black}{76.8}}^{\tiny\color{red}\textbf{-1.5}}$&\color{black}{76.4}&\color{black}{78.3}&${\color{black}{75.6}}^{\tiny\color{red}\textbf{-2.7}}$\\
SUN397 & \color{black}{62.5}&${\color{black}{67.0}}^{\tiny\color{ForestGreen}\textbf{+4.5}}$&${\color{black}{68.5}}^{\tiny\color{ForestGreen}\textbf{+6.0}}$&\color{black}{64.4}&${\color{black}{69.6}}^{\tiny\color{ForestGreen}\textbf{+5.2}}$&\color{black}{69.8}&\color{black}{67.6}&${\color{black}{72.6}}^{\tiny\color{ForestGreen}\textbf{+5.0}}$\\
FGVC Aircraft & \color{black}{19.6}&${\color{black}{16.7}}^{\tiny\color{red}\textbf{-2.9}}$&${\color{black}{23.1}}^{\tiny\color{ForestGreen}\textbf{+3.5}}$&\color{black}{24.3}&${\color{black}{17.7}}^{\tiny\color{red}\textbf{-6.6}}$&\color{black}{18.5}&\color{black}{31.8}&${\color{black}{25.0}}^{\tiny\color{red}\textbf{-6.8}}$\\
Country211 & \color{black}{17.2}&${\color{black}{14.8}}^{\tiny\color{red}\textbf{-2.4}}$&${\color{black}{16.5}}^{\tiny\color{red}\textbf{-0.7}}$&\color{black}{22.8}&${\color{black}{18.1}}^{\tiny\color{red}\textbf{-4.7}}$&\color{black}{18.9}&\color{black}{31.9}&${\color{black}{23.0}}^{\tiny\color{red}\textbf{-8.9}}$\\
DTD & \color{black}{44.3}&${\color{black}{54.6}}^{\tiny\color{ForestGreen}\textbf{+10.3}}$&${\color{black}{53.9}}^{\tiny\color{ForestGreen}\textbf{+9.6}}$&\color{black}{44.9}&${\color{black}{51.3}}^{\tiny\color{ForestGreen}\textbf{+6.4}}$&\color{black}{55.5}&\color{black}{55.3}&${\color{black}{60.5}}^{\tiny\color{ForestGreen}\textbf{+5.2}}$\\
GTSRB & \color{black}{32.6}&${\color{black}{42.0}}^{\tiny\color{ForestGreen}\textbf{+9.4}}$&${\color{black}{36.5}}^{\tiny\color{ForestGreen}\textbf{+3.9}}$&\color{black}{43.3}&${\color{black}{43.5}}^{\tiny\color{ForestGreen}\textbf{+0.2}}$&\color{black}{49.4}&\color{black}{50.6}&${\color{black}{49.9}}^{\tiny\color{red}\textbf{-0.7}}$\\
STL10 & \color{black}{97.1}&${\color{black}{95.6}}^{\tiny\color{red}\textbf{-1.5}}$&${\color{black}{96.5}}^{\tiny\color{red}\textbf{-0.6}}$&\color{black}{98.2}&${\color{black}{97.0}}^{\tiny\color{red}\textbf{-1.2}}$&\color{black}{97.0}&\color{black}{99.4}&${\color{black}{98.1}}^{\tiny\color{red}\textbf{-1.3}}$\\
Retino & \color{black}{45.5}&${\color{black}{24.2}}^{\tiny\color{red}\textbf{-21.3}}$&${\color{black}{19.1}}^{\tiny\color{red}\textbf{-26.4}}$&\color{black}{3.3}&${\color{black}{7.4}}^{\tiny\color{ForestGreen}\textbf{+4.1}}$&\color{black}{9.2}&\color{black}{73.3}&${\color{black}{6.0}}^{\tiny\color{red}\textbf{-67.3}}$\\
EuroSAT & \color{black}{50.4}&${\color{black}{51.5}}^{\tiny\color{ForestGreen}\textbf{+1.1}}$&${\color{black}{50.3}}^{\tiny\color{red}\textbf{-0.1}}$&\color{black}{55.9}&${\color{black}{50.3}}^{\tiny\color{red}\textbf{-5.6}}$&\color{black}{58.2}&\color{black}{62.6}&${\color{black}{62.3}}^{\tiny\color{red}\textbf{-0.3}}$\\
RESISC45 & \color{black}{53.6}&${\color{black}{54.5}}^{\tiny\color{ForestGreen}\textbf{+0.9}}$&${\color{black}{61.9}}^{\tiny\color{ForestGreen}\textbf{+8.3}}$&\color{black}{58.2}&${\color{black}{58.5}}^{\tiny\color{ForestGreen}\textbf{+0.3}}$&\color{black}{61.4}&\color{black}{63.4}&${\color{black}{67.4}}^{\tiny\color{ForestGreen}\textbf{+4.0}}$\\
PCAM & \color{black}{62.3}&${\color{black}{55.9}}^{\tiny\color{red}\textbf{-6.4}}$&${\color{black}{50.7}}^{\tiny\color{red}\textbf{-11.6}}$&\color{black}{50.7}&${\color{black}{59.6}}^{\tiny\color{ForestGreen}\textbf{+8.9}}$&\color{black}{55.2}&\color{black}{52.0}&${\color{black}{49.6}}^{\tiny\color{red}\textbf{-2.4}}$\\
CLEVR Counts & \color{black}{23.2}&${\color{black}{16.2}}^{\tiny\color{red}\textbf{-7.0}}$&${\color{black}{19.2}}^{\tiny\color{red}\textbf{-4.0}}$&\color{black}{21.2}&${\color{black}{28.7}}^{\tiny\color{ForestGreen}\textbf{+7.5}}$&\color{black}{23.9}&\color{black}{19.4}&${\color{black}{24.2}}^{\tiny\color{ForestGreen}\textbf{+4.8}}$\\
CLEVR Dist & \color{black}{16.3}&${\color{black}{15.9}}^{\tiny\color{red}\textbf{-0.4}}$&${\color{black}{16.8}}^{\tiny\color{ForestGreen}\textbf{+0.5}}$&\color{black}{15.8}&${\color{black}{24.5}}^{\tiny\color{ForestGreen}\textbf{+8.7}}$&\color{black}{15.9}&\color{black}{16.1}&${\color{black}{14.9}}^{\tiny\color{red}\textbf{-1.2}}$\\
DSPRITES Orient & \color{black}{2.4}&${\color{black}{1.9}}^{\tiny\color{red}\textbf{-0.5}}$&${\color{black}{2.3}}^{\tiny\color{red}\textbf{-0.1}}$&\color{black}{2.3}&${\color{black}{2.9}}^{\tiny\color{ForestGreen}\textbf{+0.6}}$&\color{black}{2.7}&\color{black}{2.3}&${\color{black}{2.6}}^{\tiny\color{ForestGreen}\textbf{+0.3}}$\\
DSPRITES pos & \color{black}{3.6}&${\color{black}{2.8}}^{\tiny\color{red}\textbf{-0.8}}$&${\color{black}{3.1}}^{\tiny\color{red}\textbf{-0.5}}$&\color{black}{3.0}&${\color{black}{3.2}}^{\tiny\color{ForestGreen}\textbf{+0.2}}$&\color{black}{4.3}&\color{black}{3.2}&${\color{black}{3.0}}^{\tiny\color{red}\textbf{-0.2}}$\\
SmallNORB Elv & \color{black}{12.7}&${\color{black}{9.9}}^{\tiny\color{red}\textbf{-2.8}}$&${\color{black}{11.0}}^{\tiny\color{red}\textbf{-1.7}}$&\color{black}{12.2}&${\color{black}{10.0}}^{\tiny\color{red}\textbf{-2.2}}$&\color{black}{11.0}&\color{black}{11.5}&${\color{black}{11.0}}^{\tiny\color{red}\textbf{-0.5}}$\\
SmallNORB Azim & \color{black}{6.1}&${\color{black}{4.5}}^{\tiny\color{red}\textbf{-1.6}}$&${\color{black}{5.2}}^{\tiny\color{red}\textbf{-0.9}}$&\color{black}{5.2}&${\color{black}{6.0}}^{\tiny\color{ForestGreen}\textbf{+0.8}}$&\color{black}{5.5}&\color{black}{4.5}&${\color{black}{5.3}}^{\tiny\color{ForestGreen}\textbf{+0.8}}$\\
DMLAB & \color{black}{19.3}&${\color{black}{17.3}}^{\tiny\color{red}\textbf{-2.0}}$&${\color{black}{18.9}}^{\tiny\color{red}\textbf{-0.4}}$&\color{black}{15.5}&${\color{black}{15.1}}^{\tiny\color{red}\textbf{-0.4}}$&\color{black}{14.8}&\color{black}{16.3}&${\color{black}{18.7}}^{\tiny\color{ForestGreen}\textbf{+2.4}}$\\
KITTI Dist & \color{black}{27.4}&${\color{black}{28.8}}^{\tiny\color{ForestGreen}\textbf{+1.4}}$&${\color{black}{17.6}}^{\tiny\color{red}\textbf{-9.8}}$&\color{black}{26.4}&${\color{black}{18.1}}^{\tiny\color{red}\textbf{-8.3}}$&\color{black}{28.1}&\color{black}{21.8}&${\color{black}{20.1}}^{\tiny\color{red}\textbf{-1.7}}$\\\midrule
\bf{VTAB+(Avg.)} & \color{black}{45.4}&${\color{black}{45.6}}^{\tiny\color{ForestGreen}\textbf{+0.2}}$&${\color{black}{47.9}}^{\tiny\color{ForestGreen}\textbf{+2.5}}$&\color{black}{47.5}&${\color{black}{48.3}}^{\tiny\color{ForestGreen}\textbf{+0.8}}$&\color{black}{49.2}&\color{black}{55.7}&${\color{black}{51.8}}^{\tiny\color{red}\textbf{-3.9}}$\\
\bottomrule\end{tabular}

\caption{Comparison between CLIP models trained on LAION (400M, 2B) and the original CLIP models \cite{radford2021learning} trained on OpenAI's WebImageText (WIT) dataset. We show zero-shot top-1 classification accuracy (\%) on the 35 datasets that are part of VTAB+. We highlight the difference ({\color{ForestGreen}+}/{\color{red}-}) between LAION models and original CLIP WIT models for each model size (except B/16+, for which there is no CLIP WIT checkpoint).
}
\label{table:zeroshot_detailed}
\end{table}

\begin{table}[p]
\rowcolors{}{}{}
\rowcolors{2}{light-light-gray}{white}
\begin{tabular}{ll|cccccc}\toprule
                          &            & \multicolumn{6}{c}{Flickr30K (1K test set)}   \\
Model & \multicolumn{1}{c|}{Pre-training} & \multicolumn{3}{c}{Image → Text} & \multicolumn{3}{c}{Text → Image} \\ \hline
                          &            & R@1   & R@5   & R@10  & R@1   & R@5   & R@10  \\
 \multirow{1}{*}{ViT-B/32}& CLIP WIT   & 77.5 & 94.7 & 98.2 & 58.8 & 83.3 & 89.7 \\
                          & LAION-400M & 78.9 & 94.0 & 97.1 & 61.7 & 85.5 & 90.9 \\
 & LAION-2B-en & 84.3 & 96.3 & 98.4 & 66.3 & 88.2 & 93.2 \\\midrule
 \multirow{1}{*}{ViT-B/16}& CLIP WIT   & 81.9 & 96.2 & 98.8 & 81.9 & 96.2 & 98.8 \\
 & LAION-400M & 83.3 & 96.8 & 98.5 & 65.5 & 88.3 & 93.0 \\\midrule
ViT-B/16+                 & LAION-400M & 86.5 & 97.1 & 98.8 & 68.0 & 88.9 & 94.0 \\\midrule
\multirow{1}{*}{ViT-L/14} & CLIP WIT   & 85.1 & 97.3 & 99.0 & 65.2 & 87.3 & 92.0 \\
& LAION-400M & 87.6 & 97.7 & 99.5 & 70.3 & 90.9 & 94.6\\\bottomrule
\caption{CLIP Zero-Shot retrieval results on the Flickr30K test set. We show retrieval performance at 1, 5, and 10 samples for both image to text and text to image.}
\label{table:flickr}
\end{tabular}
\end{table}

\begin{table}[p]
\rowcolors{2}{light-light-gray}{white}
\begin{tabular}{ll|cccccc}\toprule
                          &            & \multicolumn{6}{c}{MSCOCO (5K test set)}      \\
Model & \multicolumn{1}{c|}{Pre-training} & \multicolumn{3}{c}{Image → Text} & \multicolumn{3}{c}{Text → Image} \\ \hline
                          &            & R@1   & R@5   & R@10  & R@1   & R@5   & R@10  \\
 \multirow{1}{*}{ViT-B/32}& CLIP WIT   & 50.0 & 75.0 & 83.3 & 30.4 & 54.8 & 66.1 \\
                          & LAION-400M & 53.5 & 77.2 & 85.4 & 34.9 & 60.3 & 71.1 \\
 & LAION-2B-en & 56.4 & 79.6 & 87.4 & 38.7 & 64.1 & 74.4 \\\midrule
 \multirow{1}{*}{ViT-B/16}& CLIP WIT   & 51.7 & 76.8 & 84.3 & 32.7 & 57.8 & 68.2 \\
& LAION-400M & 56.5 & 80.4 & 87.3 & 37.9 & 63.2 & 73.3 \\\midrule
ViT-B/16+                 & LAION-400M & 58.6 & 81.6 & 88.4 & 40.0 & 65.5 & 75.1 \\\midrule
 \multirow{1}{*}{ViT-L/14}& CLIP WIT   & 56.0 & 79.5 & 86.9 & 35.3 & 60.0 & 70.2 \\
  & LAION-400M & 59.3 & 81.9 & 89.0 & 42.0 & 67.2 & 76.6\\\bottomrule
\caption{CLIP Zero-Shot retrieval results on the MSCOCO test set. We show retrieval performance at 1, 5, and 10 samples for both image to text and text to image.}
\label{table:mscoco}
\end{tabular}
\end{table}

\paragraph{Diabetic Retinopathy} We observe a large variation of performance on Diabetic Retinopathy~\cite{diabetic_retino} (noted Retino). Accuracy goes from 3\% to 73.3\% for CLIP WIT models, and from 7.4\% to 24.2\% for LAION models. Additionally, the difference  between CLIP WIT and LAION models goes up to 67.3\% (on L/14). After investigating, we found that on low accuracy models, performance on the majority class is very low (e.g., for ViT-B/16 LAION model, recall was 3.4\% on the majority class), and given that the dataset is highly imbalanced (majority class constitutes 74\% of the samples), accuracy is affected heavily.  A possible reason for low performance could be the prompts that were used, thus tuning the prompts could alleviate the problem. 
We re-evaluated the models using mean per-class recall, and found that the performances are less disparate, with a maximum difference between CLIP WIT models and LAION models of 2.1\%. Overall, the results  remain quite low, best mean per-class recall  was 25.4\%, obtained with ViT-B/32 trained on LAION-400M.

\paragraph{Structured tasks} Similarly to~\cite{zhai2021lit}, we observe low accuracy on VTAB's structured tasks~\cite{zhai2019large} (CLEVR, DSPRITES, SmallNORB, DMLAB, KITTI) which involve counting, depth prediction, or position/angle prediction. Finding ways to improve accuracy on those tasks is an open research question~\cite{zhai2021lit} that would be interesting to investigate in future work.

\paragraph{Retrieval} We observe consistent improvements of LAION models over CLIP WIT models on MSCOCO 5K test set (Tab.~\ref{table:mscoco}) across all metrics and model sizes.
On Flickr30k (Tab~\ref{table:flickr}), we observe similar or better results with LAION models, with the exception of image retrieval  on ViT-B/16  where CLIP WIT model
is better. 
It would be interesting to investigate why LAION models have an advantage, and whether the advantage is more general or  specific to the datasets that are 
considered in this work. Overall, we obtain better results than the best reported results in~\cite{radford2021learning}, e.g. on MSCOCO  text retrieval we obtain 59.3\% vs 58.4\% for CLIP WIT, and on image retrieval we obtain 42\% vs 37.8\% for CLIP WIT, both evaluated using the R@1 metric.

%% file: sections/experiments_generative_appendix.tex
Here we provide overview about training experiments that were performed with generative models, GLIDE and Stable Diffusion, using subsets of LAION-5B. 

\subsection{GLIDE}
\label{sec:appendix_experiments_GLIDE}

OpenAI released checkpoints for the GLIDE \citep{GLIDE} architecture to the public, but only released checkpoints trained on a filtered dataset removing hate-symbols and humans. These models can do a lot, but are incapable of generating imagery of humans. To evaluate the LAION dataset and its generalization capabilities, we aim to re-introduce the ability to generate imagery of humans into these checkpoints by finetuning them on LAION-5B.

We finetune the released GLIDE 64 pixel base (filtered) checkpoint from OpenAI on LAION-5B. For upscaling from 64x64 images to 256x256 images, we use the unmodified weights from OpenAI GLIDE-upsample-filtered. During training, captions were randomly replaced with the unconditional token 20\% of the time. All code and checkpoints are provided in our GitHub repository
\footnote{\url{https://github.com/LAION-AI/laionide}}.

We finetune LAIONIDE-v1 first, using an NVIDIA RTX 2070 Super GPU. Due to the 8GB VRAM constrain posed by the RTX 2070, we only use a batch size of 1. This initial checkpoint is provided as LAIONIDE-v1. 

To accelerate training, LAIONIDE-v2 is finetuned from LAIONIDE-v1 using an 8xA100 pod from Stability. LAIONIDE-v2 sees roughly 25 million shuffled text-image pairs from LAION-2B. Some data is filtered during finetuning: if a text-image pair’s ‘nsfw’ metadata has a value of 'NSFW' or 'LIKELY', we remove the sample. We remove any pairs where the language code is not ‘en’, to focus the model on english. We remove any images with an aspect ratio greater than 1.3 or less than 0.8. We remove all images where the smallest side is less than 256 pixels in length. Finally, we perform a sub-string search against a list of common slurs, and remove captions containing some slurs, although this is far from comprehensive.

To reduce the number of watermarks output by LAIONIDE-v2, we finetune to create LAIONIDE-v3. It sees roughly 1 million text-image pairs from a shuffled mixture of datasets: COCO 2017's training set (MS-COCO), Visual Genome, Open Images “Localized Annotations” and LAION-5B \citep{lin2014microsoft, krishna2017visual,OpenImages,OpenImagesLocNarr}. We find this reduces the number of watermarks output compared to LAIONIDE-V2 during manual analysis.

To improve inference time we make use of the pseudo linear multi-step diffusion sampling method from \citet{liu2022pseudo} as implemented by Katherine Crowson.

\begin{figure}
    \centering
    \includegraphics[width=\linewidth]{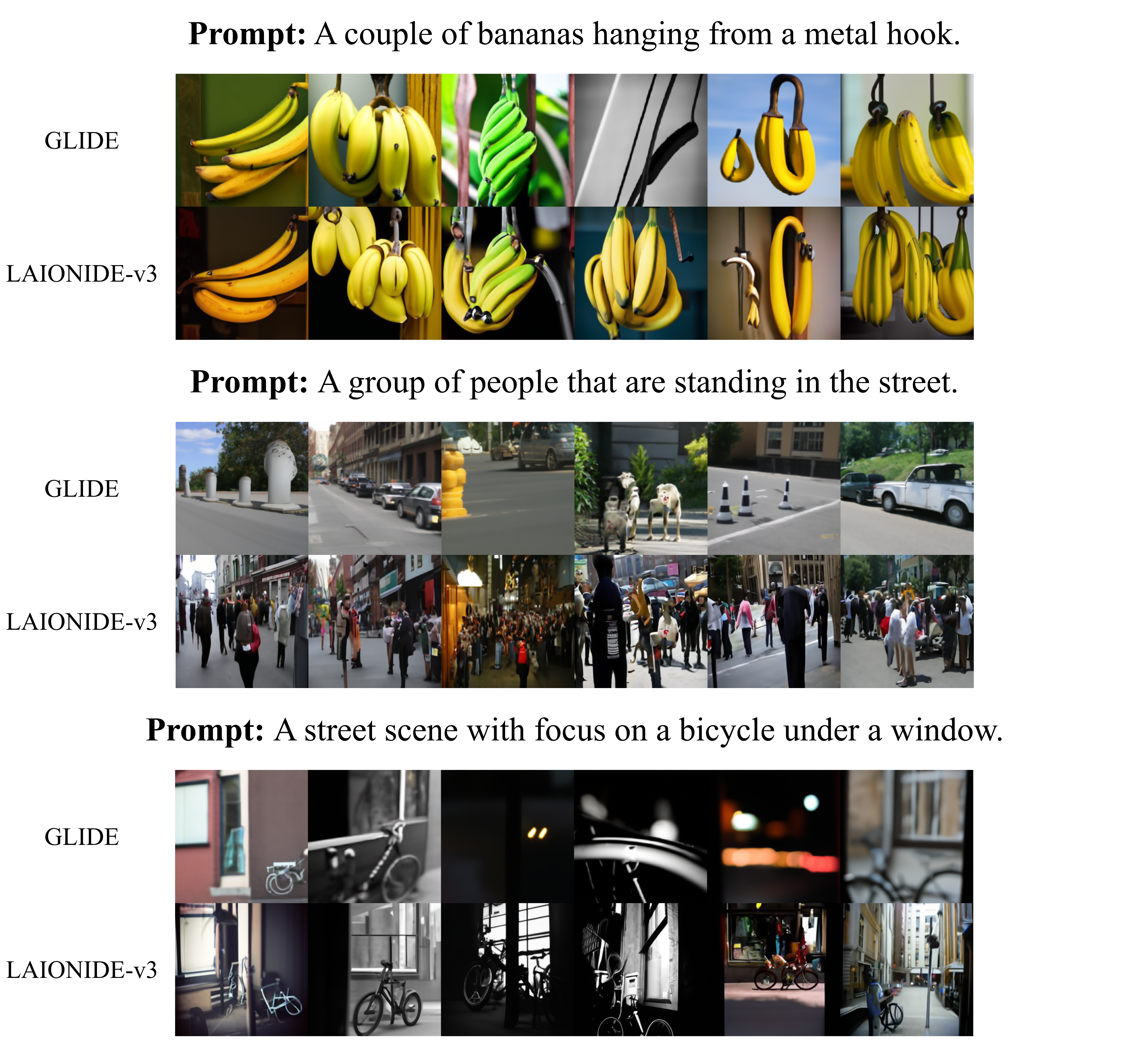}
    \caption{\textbf{Comparison of GLIDE and LAIONIDE-v3 Generations.} We compare the output of GLIDE and our LAIONIDE-v3 across three different prompts. The top row of each section depicts GLIDE's results, while the bottom row depicts LAIONIDE-v3's resutls.} 
    \label{fig:glide}
\end{figure}

We compare some evaluations from OpenAI’s released filtered checkpoint and the one we train. Those can be found at the following link: \url{https://wandb.ai/afiaka87/glide_compare/reports/laionide-v3-benchmark--VmlldzoxNTg3MTkz}



\subsection{Stable Diffusion}
\label{sec:appendix_experiments_stable_diffusion}

Stable Diffusion is a generative latent diffusion model trained on various LAION-5B subsets:

\begin{itemize}
    \item 237,000 steps at 256x256 on LAION-2B-en
    \item 194,000 steps at 512x512 on laion-high-resolution
    \item 515,000 steps at 512x512 on laion-improved-aesthetics
    \item 390,000 steps at 512x512 on laion-improved-aesthetics with 10\% dropping of the text conditioning
\end{itemize}

\begin{figure}[h!]
    \centering
    \includegraphics[width=0.5\textwidth]{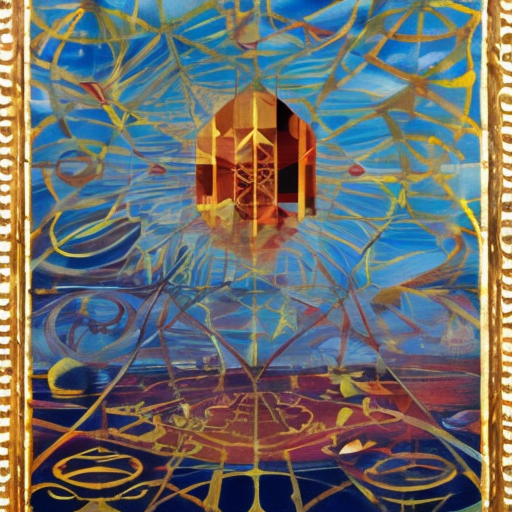}
    \caption{\centering\textbf{"The sigil of water by Gerardo Dottori, oil on canvas"}\newline Generated by Stable Diffusion}
    \label{fig:sd_generated_1}
\end{figure}

\begin{figure}[h!]
    \centering
    \includegraphics[width=0.5\textwidth]{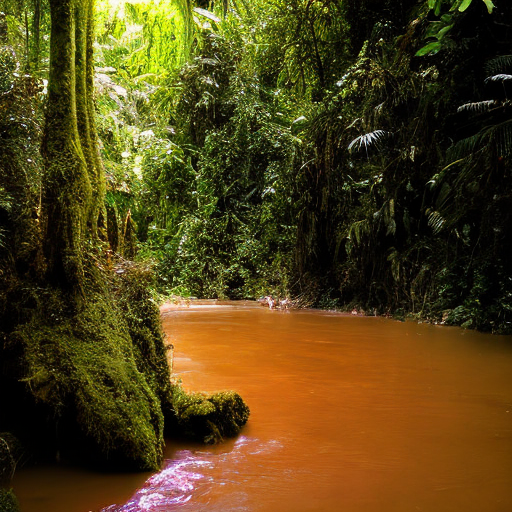}
    \caption{\centering\textbf{"A wide river in the jungle, Provia, Velvia"}\newline Generated by Stable Diffusion}
    \label{fig:sd_generated_2}
\end{figure}

Here we show representative generated samples for an artistic (Fig. \ref{fig:sd_generated_1}) and a photorealistic (Fig. \ref{fig:sd_generated_2}) image. For more technical details, we refer to the Stable Diffusion github repository\footnote{\url{https://github.com/CompVis/stable-diffusion/}}.

%% file: sections/safety_appendix.tex
\subsection{Privacy}
\label{sec:appendix_privacy}

As any other dataset of links obtained from Common Crawl that gathers content from publicly available Internet, LAION-5B can contain links to images with personal information, like to photos of faces, medical images or other personal related content. Tools like CLIP retrieval (see Appendix Section \ref{sec:appendix_ui_search} for more details) provided by LAION make it possible for the users to find out by text or image prompt whether any of the links crawled for LAION-5B point to their personal data and if yes, where on the public internet the corresponding data is hosted. Thus, for the first time, the broad public can take a look inside of a typical large-scale crawled dataset and become aware of the possible content of datasets that can be used for model training. As most of institutions and companies use same crawling procedures to obtain their closed datasets, we thus also hope to increase awareness for the risks which publicly available data can be used and exploited by third parties who do not disclose their data collection and application procedures. At the same time, researchers can access LAION-5B to study privacy related issues in such data and develop measures that increase safety of applications arising from training models on data crawled from public internet.

As LAION tools empower people to discover problematic personal or copyrighted content available in the public internet, the users can also initiate procedures of removing corresponding images from the public internet by contacting the responsible host providers that have published those images following the links provided in LAION-5B. In addition, we also provide a contact form on our website \footnote{\href{https://laion.ai/dataset-requests/}{https://laion.ai/dataset-requests/}} where requests for removal or blacklisting of the corresponding links from LAION-5B can be processed.

Further, to mitigate privacy concerns, there exist methods that allow personal human attributes like faces to be obfuscated~\citep{yang2022imagenetfaces} or generated~\citep{maximov2020ciagan} and thus made anonymous, without hurting the quality and richness of learned representations. Especially generation based methods can be applied to open data like LAION-5B to create training datasets that do not contain any private facial data, while still allowing to learn proper face representation during training. This line of work is currently in progress in LAION community.

\subsection{Potential Biases Induced by CLIP Filtering}
\label{sec:appendix_clip_filtering}

\textbf{Unknown initial dataset.}
The CLIP model in itself introduces a bias, which cannot be trivially assessed, as the underlying dataset on which the model was trained is not openly accessible. With the release of a large openly accessible image-text dataset, we offer a starting point in the open auditing of contrastive image-text models like CLIP.

\textbf{Selection heuristic based on cosine similarity.}
As noted by \cite{birhane2021multimodal}, cosine similarity is only a heuristic that also may lead to suboptimal guidance for dataset filtering. The work showed examples in which captions with malignant descriptions obtain a higher similarity over a benign description. During CLIP's training, the cosine similarity only acted as a logit to represent the likelihood of a given image-text pairing. It fails to encapsulate the nuance and rich semantic and contextual meaning that the image or language might contain. By using cosine similarity as a ground for filtering, the dataset might exacerbate those biases already contained by CLIP.

%% file: sections/author_contributions.tex

\begin{itemize}
\item \textbf{Christoph Schuhmann}: He led this project and built POCs for most of its components including clip filtering, the safety model, the watermark model and the BLIP inference tuning project.
\item \textbf{Richard Vencu}: System architecture and download script optimizations, GPU assisted filtering. Set up the AWS infrastructure.
\item \textbf{Romain Beaumont}: Guidance on scaling for the Common Crawl filtering pipeline. Built and ran the dataset preparation pipeline: pyspark deduplication job, img2dataset, CLIP inference, autofaiss, safety tags.
\item \textbf{Clayton Mullis}: DALLE-pytorch training/analysis, WDS filtering, trained generative models (LAIONIDE) using LAION-5B.
\item \textbf{Ludwig Schmidt}: Provided advice on experiment design, scaling, ethical and social content, and paper writing.
\item \textbf{Jenia Jitsev}: scientific organization \& manuscript writing, ethical and social content, experiments planning and design, compute and storage resource acquisition, general supervision.
\item \textbf{Robert Kaczmarczyk}: Established WDS architecture, performed DALL-E training runs, balancing calculation, sample (NSFW, watermark, caption quality) annotation, manuscript writing coordination, supervision and revision.
\item \textbf{Theo Coombes}: He was one of our first contributors \& build the first versions of our worker swarm system. Without his enthusiasm this project might never have taken off.
\item \textbf{Aarush Katta}: Trained the watermark model.
\item \textbf{Cade Gordon}: Ran distributed inference for the watermark tags, trained the CLIP models on JUWELS Booster, and led the paper writing.
\item \textbf{Mehdi Cherti}: Evaluated the CLIP-B/32, B/16, B/16+ and L/14 model, performed debugging of distributed training, executed experiments on JUWELS Booster, performed results collection, distillation and analysis, manuscript writing.
\item \textbf{Ross Wightman}: Ross debugged \& trained the CLIP-B/32, B/16, B/16+ and L/14 model and executed experiments on JUWELS Booster.
\item \textbf{Katherine Crowson}: Contributed to development of latent diffusion and stable diffusion. Fine-tuned generative models on subsets of LAION-5B.
\item \textbf{Patrick Schramowski}: Patrick helped with NSFW and otherwise inappropriate content tagging. Further, he wrote the corresponding parts as well as the ethical and social content.
\item \textbf{Srivatsa Kundurthy}: Co-wrote the datasheet, researched usage cases \& related works, trained face classifier and developed visualizations.
\item \textbf{Mitchell Wortsman} Initially created openCLIP, provided insights on scaling, performed experiments evaluating few-shot fine-tuning performance and robustness on ImageNet and other downstream datasets
\end{itemize}


%% file: sections/acknowledgments_appendix.tex
We want to thank our open community for their continuous efforts for openly available datasets and models. Without the broad support from the community, especially in the early crawling days with decentralized compute support, this project would not have been possible.

Moreover, the following organizations and persons contributed to this project:

\begin{itemize}
\item \textbf{Aran Komatsuzaki}: He led the initial crawling@home image-text-pair dataset building project (the predecessor of LAION-400M).

\item \textbf{Andreas Köpf}: He conducted the hyperparameter search for the inference strategies with the BLIP image-captioning model.

\item \textbf{Bokai Yu}: Accomplished most of the work to make the knn index building tool autofaiss work in a distributed setting.

\item \textbf{John David Pressman}: Provided aestethic dataset for creating aestethic LAION subset to fine-tune GLIDE.

\item \textbf{Natalie Parde} Assisted in manuscript revisions.

\item \textbf{Gabriel Ilharco} Initially created OpenCLIP and gave valuable insights on scaling.

\item \textbf{Fredde Frallan} Provided zero-shot retrieval results.

\item \textbf{Hugging Face}: provided financial and computing support, helped hosting LAION-5B as well as related subsets.

\item \textbf{Emad Mostaque (Stability AI)}: provided financial and computation support for open-source datasets and models.


\end{itemize}